\documentclass{article}

\usepackage{arxiv}

\usepackage[utf8]{inputenc} 
\usepackage[T1]{fontenc}    
\usepackage{hyperref}       
\usepackage{url}            
\usepackage{booktabs}       
\usepackage{amsfonts}       
\usepackage{nicefrac}       
\usepackage{microtype}      
\usepackage{lipsum}		
\usepackage{graphicx}
\usepackage{natbib}
\usepackage{nameref}
\usepackage{float}

\usepackage{pifont}    

\newcommand{\xmark}{\ding{55}}  

\usepackage{tabularx}
\newcolumntype{C}{>{\centering\arraybackslash}X}
\newcolumntype{L}{>{\raggedright\arraybackslash}X}

\usepackage{doi}
\usepackage[table]{xcolor} 
\usepackage{hyperref} 
\definecolor{darkblue}{rgb}{0.0, 0.0, 0.55} 
\definecolor{darkred}{rgb}{0.55, 0.0, 0.0} 
\definecolor{myblue}{RGB}{82, 140, 182}

\hypersetup{
    colorlinks=true, 
    citecolor=darkgreen, 
    linkcolor=darkred, 
    urlcolor=darkred 
}

\definecolor{green}{HTML}{1cc650}
\definecolor{darkergreen}{HTML}{006400}
\definecolor{darkblue}{rgb}{0.0, 0.0, 0.55} 
\definecolor{darkred}{rgb}{0.55, 0.0, 0.0} 

\usepackage{tcolorbox}

\newtcolorbox{myinfobox}[2][]
{
  colframe = darkergreen!50,
  colback  = mygreen2!35,
  #1, 
}

\newtcolorbox{myremark}{
  colback=mygreen!17, 
  colframe=mygreen!70!black, 
  boxrule=0pt, 
  leftrule=2pt, 
  rightrule=2pt, 
  boxsep=5pt, 
  arc=0pt, 
  left=5pt, 
  right=5pt, 
  top=0pt, 
  bottom=0pt 
}

\newtcolorbox{myremark_grey}{
  colback=mygrey!6, 
  colframe=mygrey!85!black, 
  boxrule=0pt, 
  leftrule=2pt, 
  rightrule=2pt, 
  boxsep=5pt, 
  arc=0pt, 
  left=5pt, 
  right=5pt, 
  top=0pt, 
  bottom=0pt 
}

\pdfoutput=1
\usepackage{times}
\usepackage{latexsym}
\usepackage{fontawesome5}

\newcommand{\Generation}{\textbf{G} \xspace}
\newcommand{\Classification}{\textbf{C} \xspace}
\newcommand{\Recommendation}{\textbf{R} \xspace}

\usepackage[T1]{fontenc}
\usepackage[utf8]{inputenc}
\usepackage{microtype}
\usepackage{inconsolata}
\usepackage{graphicx}
\usepackage{url}
\usepackage{pifont}
\usepackage[utf8]{inputenc} 
\usepackage[T1]{fontenc}    
\usepackage{hyperref}       
\usepackage{url}            
\usepackage{booktabs}       
\usepackage{amsfonts}       
\usepackage{nicefrac}       
\usepackage{microtype}      
\usepackage{xcolor}         
\usepackage{amsmath}
\usepackage{amsthm}
\usepackage{booktabs}
\usepackage{siunitx}
\usepackage{makecell}
\usepackage{fancyvrb}
\usepackage[edges]{forest}

\definecolor{hiddendraw}{RGB}{20,68,106}
\definecolor{hidden-pink}{RGB}{255,245,247}
\definecolor{hidden-red}{RGB}{180,0,0}
\definecolor{output-black}{RGB}{0,0,0}

\definecolor{output-black}{RGB}{0,0,0}
\definecolor{output-white}{RGB}{255,255,255}
\definecolor{myorange}{RGB}{255,208,153}
\definecolor{mygreen}{RGB}{166,207,152}
\definecolor{mygreen2}{RGB}{229,236,178}
\definecolor{forestgreen}{RGB}{34,139,34}

\definecolor{darkgreen}{RGB}{24, 119, 24}

\usepackage{bibentry}

\usepackage{verbatim}
\usepackage{multirow}
\usepackage{xspace}

\usepackage[ruled,boxed,commentsnumbered]{algorithm2e}
\usepackage{algorithmicx}
\usepackage{algpseudocode}
\usepackage{lineno}

\usepackage{colortbl}
\usepackage[table]{xcolor}

\usepackage{subcaption}
\usepackage{tabularx}
\usepackage{xcolor}
\usepackage{enumitem}
\usepackage{cases}

\usepackage{wrapfig}
\usepackage{graphicx}
\usepackage{svg}
\usepackage{bm}
\usepackage{tocloft} 
\usepackage{amsfonts}
\usepackage{pifont}
\usepackage{utfsym}
\usepackage{mathtools}

\newcommand{\textbit}[1]{\textcolor{black}{\textit{#1}}}

\usepackage[table]{xcolor}

\usepackage{threeparttable}
\definecolor{melon}{HTML}{F89E7B}
\definecolor{lightorange}{HTML}{FBB982}
\definecolor{userblack}{HTML}{41B0E4}
\definecolor{cerulean}{HTML}{00A2E3}

\usepackage{titletoc}
\usepackage{minitoc}

\definecolor{mygrey}{RGB}{128,128,128}
\definecolor{verylightgray}{RGB}{240, 240, 240}

\usepackage{tcolorbox}
\usepackage{xcolor}
\newtcolorbox{warningbox}[2][]
{
  colframe = red!25,
  colback  = red!10,
  coltitle = red!20!black,
  title    = #2,
  #1,
}

\newtcolorbox{hintbox}[2][]
{
  colframe = green!25,
  colback  = green!10,
  coltitle = green!20!black,
  #1,
}

\newtcolorbox{citebox}[2][]
{
  colframe = mygrey!55,
  colback  = mygrey!10,
  #1, 
}

\definecolor{titlebgcolor}{RGB}{82, 140, 182}
\definecolor{textbgcolor}{RGB}{219, 232, 241}

\newtcolorbox{infobox}[2][]
{
  colframe = titlebgcolor!60,
  colback  = textbgcolor!29,
  #1, 
}

\newtcolorbox{methbox}[2][]{
  colframe = titlebgcolor!78,
  colback  = textbgcolor!29,
  fonttitle = \bfseries\footnotesize,
  title    = {#2},
  boxrule  = 0.8pt,
  arc      = 4pt,      
  top      = 6pt,
  bottom   = 6pt,
  left     = 6pt,
  right    = 6pt,
  colbacktitle = titlebgcolor,
  coltitle = white,
  #1
}

\usepackage{minitoc}

\usepackage{pifont}

\def\equationautorefname~#1\null{Equation~(#1)\null}
\def\appendixautorefname~#1\null{Appendix~#1\null}
\def\subappendixautorefname~#1\null{Appendix~#1\null}
\def\sectionautorefname~#1\null{Section~#1\null}
\def\subsectionautorefname~#1\null{Section~#1\null}
\def\figureautorefname~#1\null{Figure~#1\null}
\def\tableautorefname~#1\null{Table~#1\null}
\def\observationautorefname~#1\null{Observation~#1\null}
\def\algorithmautorefname~#1\null{Algorithm~#1\null}

\usepackage{titletoc}
\usepackage{wasysym}
\usepackage{adjustbox}
\usepackage{fontawesome5}
\usepackage{pifont}
\usepackage{tocloft}

\title{A Survey of Personalized Large Language Models: \\ Progress and Future Directions}

\author{
\bf Jiahong Liu$^{\clubsuit}$\thanks{\texttt{ jiahong.liu21@gmail.com, \{zxqiu22, whyu24, mindahu21,king\}@cse.cuhk.edu.hk}
\\ \texttt{               \{lizhongyang6,daiquanyu,jiamie.zhu\}@huawei, menglin.yang@outlook.com}, dcscts@nus.edu.sg}
, Zexuan Qiu$^{\clubsuit}$, Zhongyang Li$^{\diamondsuit}$
, Quanyu Dai$^{\diamondsuit}$, Wenhao Yu$^{\clubsuit}$, Jieming Zhu$^{\diamondsuit}$, \\ 
\bf \vspace{0.3cm}
Minda Hu$^{\clubsuit}$,
Menglin Yang$^{\heartsuit}$, Tat-Seng Chua$^{\spadesuit}$, Irwin King$^{\clubsuit}$\\
$^{\clubsuit}$ The Chinese University of Hong Kong\\
$^{\diamondsuit}$ Huawei Technologies Co., Ltd\\
$^{\heartsuit}$ The Hong Kong University of Science and Technology (Guangzhou)\\
\vspace{0.2cm} 
$^{\spadesuit}$ National University of Singapore\\
\raisebox{-0.3\height}{\hspace{0.05cm}\includegraphics[width=0.57cm]{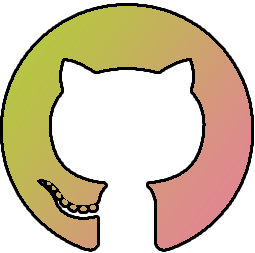}} 
\textbf{\mbox{Repository:}} \href{https://github.com/JiahongLiu21/Awesome-Personalized-Large-Language-Models}{github.com/JiahongLiu21/Awesome-Personalized-Large-Language-Models}
}

\renewcommand{\shorttitle}{A Survey of Personalized Large Language Models}

\usepackage{colortbl}

\begin{document}
\maketitle

\begin{abstract}
    Large Language Models (LLMs) excel in handling general knowledge tasks, yet they struggle with user-specific personalization, such as understanding individual emotions, writing styles, and preferences. Personalized Large Language Models (PLLMs) tackle these challenges by leveraging individual user data, such as user profiles, historical dialogues, content, and interactions, to deliver responses that are contextually relevant and tailored to each user's specific needs. This is a highly valuable research topic, as PLLMs can significantly enhance user satisfaction and have broad applications in conversational agents, recommendation systems, emotion recognition, medical assistants, and more. This survey reviews recent advancements in PLLMs from three technical perspectives: prompting for personalized context (input level), finetuning for personalized adapters (model level), and alignment for personalized preferences (objective level). To provide deeper insights, we also discuss current limitations and outline several promising directions for future research. 
\end{abstract}


\vspace{15pt}
\begin{figure*}[htbp]
    \centering
    \includegraphics[width=0.999\linewidth]{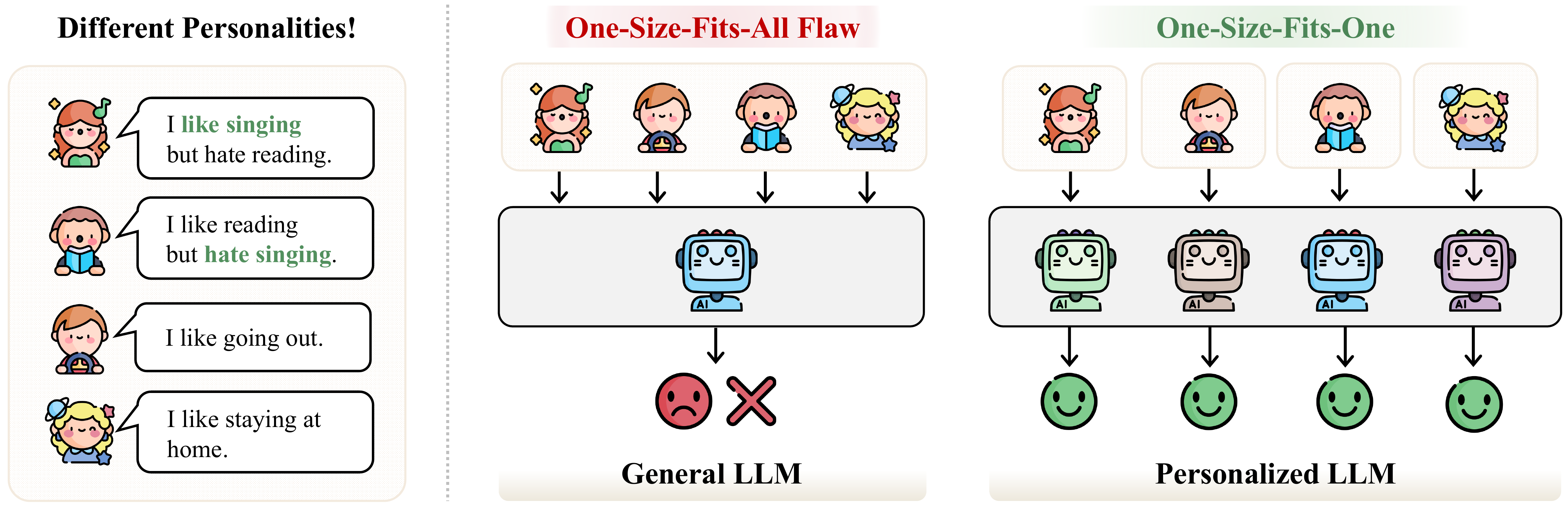}
    \caption{ Comparison of General LLM vs. Personalized LLM. Different users have different preferences. General LLMs fail to satisfy diverse needs with a one-size-fits-all approach, demonstrating the necessity for personalized LLMs.}
    \label{fig:illustarton}
\end{figure*}

\newpage

\tableofcontents

\newpage

\section{Introduction}

In recent years, substantial progress has been made in Large Language Models (LLMs) such as GPT, PaLM, LLaMA, DeepSeek, and their variants~\citep{zhao2023survey}. These models have demonstrated remarkable versatility, achieving state-of-the-art performance across various natural language processing (NLP) tasks, including question answering, logical reasoning, and machine translation~\citep{chang2024survey, hu-etal-2024-serts,zhang2024memsim,zhang2024survey, zhu2024lifelong,wang2023large,wang2024tpe,li2024personal, zhao2025towards, zhu2025soft}, with minimal task-specific adaptation.

\begin{figure}[!t]
    \centering
    \includegraphics[width=0.9\linewidth]{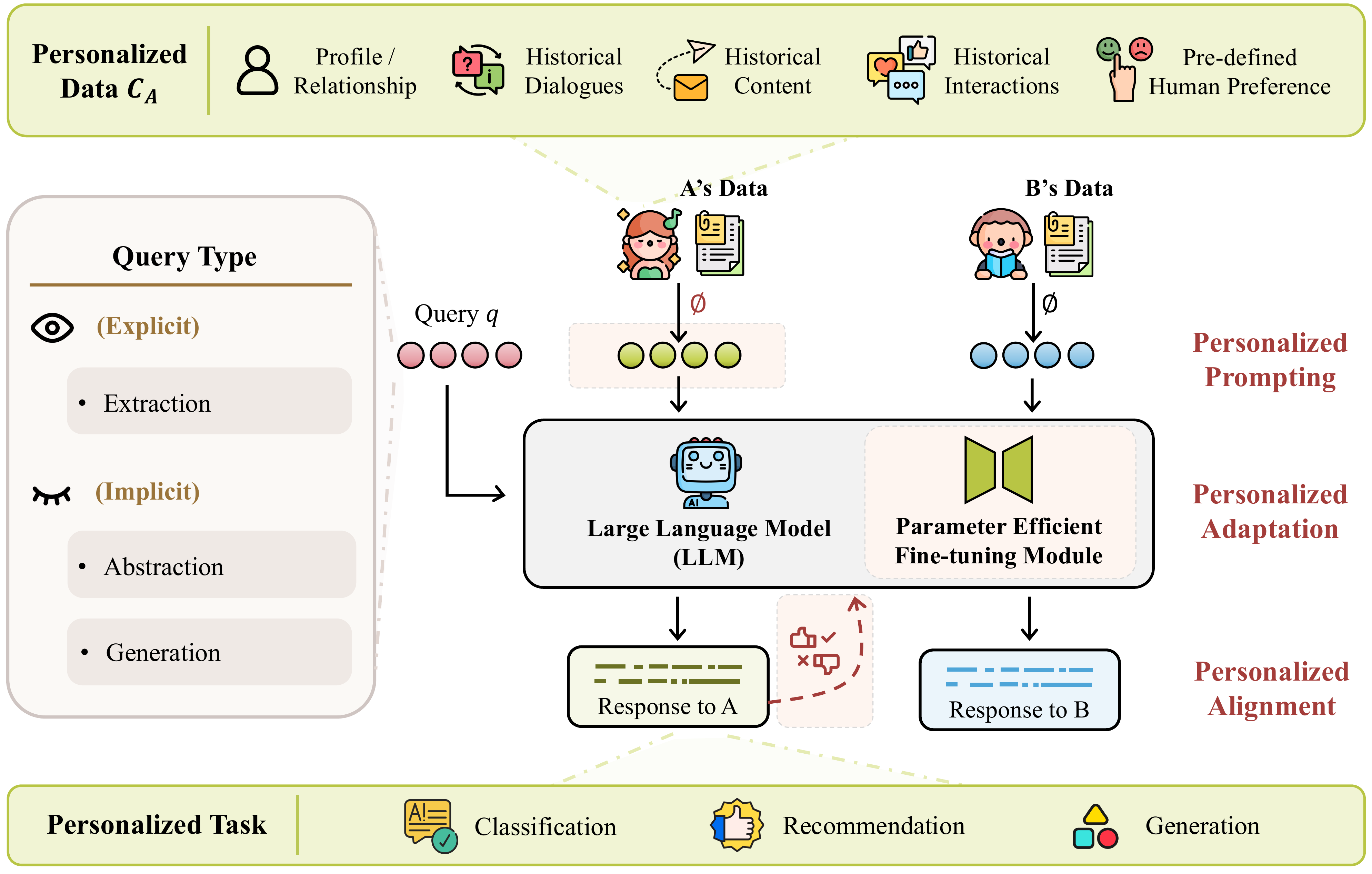}
    \caption{Illustration of PLLM techniques for generating personalized responses through three levels: prompting (input level, \autoref{sec: personalized prompting}), adaptation (model level, \autoref{sec: personalized adatation}), and alignment (objective level, \autoref{sec: personalized alignment}). \nameref{para:personalized data}}
    \label{fig:framework}
\end{figure}

\tikzstyle{leaf}=[draw=hiddendraw,
    rounded corners,minimum height=1em,
    fill=mygreen2!38,text opacity=1, align=center,
    fill opacity=.5, text=black, align=left,font=\scriptsize,
    inner xsep=3pt,
    inner ysep=1pt,
    ]
\tikzstyle{middle}=[draw=hiddendraw,
    rounded corners,minimum height=1em,
    fill=output-white!40,text opacity=1, align=center,
    fill opacity=.5,  text=black,align=center,font=\scriptsize,
    inner xsep=3pt,
    inner ysep=1pt,
    ]
\begin{figure*}[ht]
\centering
\begin{forest}
  for tree={
  forked edges,
  grow=east,
  reversed=true,
  anchor=base west,
  parent anchor=east,
  child anchor=west,
  base=middle,
  font=\scriptsize,
  rectangle,
  line width=0.7pt,
  draw=output-black,
  rounded corners,align=left,
  minimum width=2em,
    s sep=5pt,
    inner xsep=3pt,
    inner ysep=1pt,
  },
  where level=1{text width=4.5em}{},
  where level=2{text width=6em,font=\scriptsize}{},
  where level=3{font=\scriptsize}{},
  where level=4{font=\scriptsize}{},
  where level=5{font=\scriptsize}{},
  [Personalized Large Language Models \\ (PLLMs), middle, align=center, rotate=90,anchor=north,edge=black
    [Personalized Prompting \\  (Input level), middle, align=center, edge=output-black,text width=6.8em
        [Profile-Augmented \\  (\textsection\ref{subsec:PAG}), middle, text width=7.2em, align=center, edge=output-black
            [ Cue-CoT~\citep{wang2023cue}{,}
             PAG~\citep{richardson2023integrating}{,}
             ONCE~\citep{DBLP:conf/wsdm/LiuCS024}{,} \\
             Matryoshka~\citep{li2024matryoshka}{,}  
             DPL~\citep{qiu2025measuring}
             {,} 
             RewriterSlRl~\citep{li2024learning}{,}\\
            R2P~\citep{luo2025reasoning}
           , leaf, text width=25.8em, edge=output-black]
        ]
        [Retrieval-Augmented \\ (\textsection\ref{subsec:RAG}), middle, text width=7.2em, align=center, edge=output-black
            [Memory, middle, text width=2.5em, edge=output-black
                [
                MemPrompt~\citep{madaan2022memory}{,}
                ~\citep{zhang2023long}{,} \\
                MaLP~\citep{zhang2024llm}{,}
                TeachMe~\citep{dalvi2022towards}{,} \\
                MemoRAG~\citep{qian2024memorag}{,} 
                FERMI~\citep{kim2024few}
                , leaf, text width=21.5em, edge=output-black]
            ]
            [Retriever, middle, text width=2.5em, edge=output-black
                [
                IPA{,}FiD~\citep{salemi2023lamp}{,}
                MSP~\citep{zhong2022less}{,} \\
                AuthorPred~\citep{li2023teach}{,}
                PEARL~\citep{mysore2023pearl}{,}\\
                ROPG~\citep{salemi2024optimization}{,}
                HYDRA~\citep{zhuang2024hydra}{,} \\
                RECAP~\citep{liu2023recap}
                {,}
                CFRAG~\citep{shi2025retrieval}{,}
                \\
                AP-Bots~\citep{yazan2025improving}{,} RPM~\citep{kim2025llms}{,}  \\
                PersonaAgent~\citep{zhang2025personaagent}
                , leaf, text width=21.5em, edge=output-black]
            ]
        ]
        [{\color{black}Soft-Fused \\ (\textsection\ref{subsec:soft})}, middle, align=center, text width=7.2em, edge=output-black
            [
            UEM~\citep{doddapaneni2024user}{,}
            PERSOMA~\citep{hebert2024persoma}{,}\\
            REGEN~\citep{sayana2024beyond}{,}
            PeaPOD~\citep{ramos2024preference}{,} 
            PPlug~\citep{liu2024llms+}{,} \\
            User-LLM~\citep{ning2024user}{,} 
            RECAP~\citep{liu2023recap}{,} 
            ComMer~\citep{zeldes2025commer}
            , leaf, text width=25.8em, edge=output-black]
        ]
        [{\color{black} Contrastive \\ (\textsection\ref{subsec: contrastive prompting})}, middle, align=center, text width=7.2em, edge=output-black
            [
            CoS~\citep{he2024cos}{,}
            StyleVector~\citep{zhang2025personalized}
            , leaf, text width=25.8em, edge=output-black]
        ]
    ]
    [Personalized Adaptation \\  (Model level), middle, align=center, edge=output-black, text width=6.8em
        [One PEFT All Users \\ (\textsection\ref{subsec:one4all}) , middle, align=center, text width=7.2em, edge=output-black
            [
            PEFT-U~\citep{clarke2024peft}{,}
            PLoRA~\citep{DBLP:conf/aaai/ZhangWYXZ24}{,} \\
            LM-P~\citep{wozniak2024personalized}{,}        
            Review-LLM~\citep{peng2024llm}{,} \\
            MiLP~\citep{zhang2024personalized}{,}
            RecLoRA~\citep{zhu2024lifelong}{,}
            iLoRA~\citep{kongcustomizing}
            , leaf, text width=25.8em, edge=output-black]
        ]
        [One PEFT Per User \\ (\textsection\ref{subsec:one4one}), middle, align=center, text width=7.2em, edge=output-black
            [
            UserAdapter~\citep{zhong2021useradapter}{,}
            PocketLLM~\citep{peng2024pocketllm}{,}  \\
            OPPU~\citep{DBLP:conf/emnlp/Tan000Y024}{,} 
            PER-PCS~\citep{tan2024personalized}{,}
            \citep{wagner2024personalized}{,} \\
            FDLoRA~\citep{qi2024fdlora}{,} 
            HYDRA~\citep{zhuang2024hydra}{,} \\
            PROPER~\citep{zhang2025proper}
            , leaf, text width=25.8em, edge=output-black]
        ]
    ]
    [Personalized Alignment \\  (Objective level), middle, align=center, edge=output-black, text width=6.8em
        [Data Construction \\ (\textsection\ref{subsec:data}), middle, align=center, text width=7.2em, edge=output-black
            [\citep{wu2024aligning}{,}
            PLUM~\citep{magister2024way}{,}
            \citep{lee2024aligning}{,}\\
            \citep{qin2024enabling}{,}
            PRISM~\citep{kirk2024prism}{,}
        PersonalLLM~\citep{zollo2024personalllm}
            ,leaf, text width=25.8em, edge=output-black]
        ]
        [Optimization \\(\textsection\ref{subsec:optimization}), align=center, middle, text width=7.2em, edge=output-black
            [MORLHF~\citep{wu2023fine}{,}
            MODPO~\citep{zhou2023beyond}{,} \\
            Reward Soups~\citep{rame2024rewarded}{,} 
            Personalized Soups~\citep{jang2023personalized}{,} \\
            MOD~\citep{shi2024decoding}{,} 
            BiPO~\citep{cao2024personalized}{,}
            PAD~\citep{chen2024pad}{,} \\
            CIPHER~\citep{gaoaligning}{,}
            Amulet~\citep{zhangamulet}{,} PPT~\citep{lau2024personalized}{,} \\
            VPL~\citep{poddar2024personalizing}{,}
            CHAMELEON~\citep{zhang2025personalize}{,} \\
            REST-PG~\citep{salemi2025reasoning}{,}
            Drift~\citep{kim2025drift}{,} 
            RLPA~\citep{zhao2025teaching}{,} \\
            PROSE~\citep{arocaaligning}{,}
            COPE~\citep{bu2025personalized}
            ,
            ,leaf, text width= 25.8em, edge=output-black]
        ]
    ]
    [ Others, middle, edge=output-black, text width=6.8em
        [Analysis, middle, text width=7.2em, align=center, edge=output-black, align=center
            [Role of User Profile~\citep{wu2024understanding, zhao2025exploring}{,} \\
            Safety-Utility~\citep{vijjini2024exploring} {,}
            RAG vs. PEFT~\citep{salemi2024comparing}
            , leaf, text width=25.8em, edge=output-black]
        ]
        [Benchmark\\ (\textsection\ref{sec: benchmark}), middle, text width=7.2em, align=center, edge=output-black
            [LaMP~\citep{salemi2023lamp}{,}
            LongLamp~\citep{kumar2024longlamp}{,} \\ 
            ALOE~\citep{wu2024aligning} {,} 
            PGraphRAG~\citep{au2025personalized}{,} \\
            PerLTQA~\citep{du2024perltqa}{,}
            PEFT-U~\citep{clarke2024peft}{,} \\
            REGEN~\citep{sayana2024beyond} {,} 
        PersonalLLM~\citep{zollo2024personalllm}{,}  \\
            PrefEval~\citep{zhao2025llms}{,}
            LongMemEval~\citep{wulongmemeval}
            , leaf, text width=25.8em, edge=output-black]
        ]
    ]
  ]
\end{forest}
\caption{A taxonomy of PLLMs with representative examples.}
\label{fig:taxonomy_of_PLLM}
\end{figure*}

\paragraph{The Necessity of Personalized LLMs (PLLMs)} While LLMs excel in general knowledge and multi-domain reasoning, their lack of personalization creates challenges in situations where user-specific understanding is crucial. For instance, conversational agents need to adapt to a user's preferred tone and incorporate past interactions to deliver relevant, personalized responses. 
As LLMs evolve, integrating personalization capabilities has become a promising direction for advancing human-AI interaction across diverse domains such as recommendation, education, healthcare, and finance~\citep{wang2024towards, hu-etal-2024-serts,zhang2024memsim,zhang2024survey, zhu2024lifelong,wang2023large,wang2024tpe}.
Despite its promise, personalizing LLMs presents several challenges. These include efficiently representing and integrating diverse user data, addressing privacy concerns, managing long-term user memories, inferring users' implicit preferences, etc
~\citep{salemi2023lamp}. 
Moreover, achieving personalization often requires balancing accuracy and efficiency while addressing biases and maintaining fairness in the outputs.

\paragraph{Contributions} Despite growing interest, the field of PLLMs lacks a systematic review that consolidates recent advancements.
This survey aims to bridge the gap by systematically organizing existing research on PLLMs and offering insights into their methodologies and future directions. The contributions of this survey are as follows:

\begin{enumerate}[label={(\arabic*)}]
\item \textit{A systematic formulation of personalization scenarios:} We provide precise problem formulations for different personalized LLM scenarios based on query types and personalized data characteristics, establishing a unified framework that distinguishes between different personalized techniques (\autoref{sec: problem statement}).

\item \textit{A structured taxonomy:} We propose a comprehensive taxonomy, providing a technical perspective on the existing approaches to building PLLMs across three levels: input-level prompting, model-level adaptation, and objective-level alignment (\autoref{sec:taxonomy}).

\item \textit{A comprehensive review:} We systematically review state-of-the-art methods for PLLMs, analyzing fine-grained differences among the methods and their applicability to different personalization scenarios (\autoref{sec: personalized prompting}, \autoref{sec: personalized adatation}, \autoref{sec: personalized alignment}).

\item \textit{A benchmark and evaluation summarization:} We provide summarization of metrics, benchmarks tailored to different query types (extraction, abstraction, and generalization) and personalization scenarios (\autoref{sec: benchmark}).

\item \textit{Future directions:} We highlight current limitations
and outline promising avenues for future research, including multimodal personalization, edge computing, lifelong updating, trustworthiness, etc (\autoref{sec: future directions}).

\end{enumerate}

\section{Preliminary}

\subsection{Large Language Models}
Large Language Models (LLMs) generally refer to models that utilize the Transformer architecture and are equipped with billions of parameters trained on trillions of text tokens. These models have demonstrated substantial improvements in a myriad of tasks related to natural language understanding and generation, increasingly proving beneficial in assisting human activities. In this work, we mainly focus on autoregressive LLMs, which are based on two main architectures:  decoder-only models and encoder-decoder models. Encoder-decoder models such as Flan-T5~\citep{chung2022scaling} and ChatGLM~\citep{zeng2022glm} analyze input through the encoder for semantic representations,  making them effective in language understanding in addition to generation. 
Decoder-only LLMs focus on left-to-right generation by predicting the next token in a sequence, with numerous instances~\citep{brown2020language, chowdhery2022palm, touvron2023llama, guo2025deepseek} under this paradigm achieving breakthroughs in advanced capabilities.

However, these models are typically pre-trained on general-purpose data and \textbf{lack an understanding of specific user information}. As a result, they are unable to generate responses tailored to a user's unique tastes and expectations, limiting their effectiveness in personalized applications where user-specific adaptation is critical.

\subsection{Problem Statement}
\label{sec: problem statement}

Personalized Large Language Models (PLLMs) generate responses that align with the user's style and expectations, offering diverse answers to the same query for different users~\citep{clarke2024peft}.

\paragraph{Problem Formulation}

Let $\mathcal{U} = \{u_i\}_{i=1}^N$ be a finite set of users. For each $u_i \in \mathcal{U}$:
\begin{itemize}
    \item $\mathcal{Q}_i = \{q_j^{(i)}\}_{j=1}^{n_i} \subset \mathcal{Q}$, where $n_i \in \mathbb{N}$ and $\mathcal{Q}$ is the query space;
    \item $\mathcal{C}_i \subset \mathcal{C}$ is the set of $u_i$'s personalized data, where $\mathcal{C}$ denotes the space of all user personalized data. 
\end{itemize}

Let $\mathcal{Y}$ be the output space. For each user $u_i$ and query $q_j^{(i)}$, there exists a desired output $\hat{y}_j^{(i)} \in \mathcal{Y}$ that aligns with $u_i$'s preferences and expectations. 
We define a metric function $\zeta: \mathcal{Y} \times \mathcal{Y} \rightarrow \mathbb{R}_{\geq 0}$ that evaluates the quality of the LLM-generated output compared with the desired output (the summarization of metrics is shown in \autoref{subsec: metric}, \autoref{tab:metrics_specs}). Here, we generally assume $\zeta(\hat{y}_j^{(i)}, y_{j}^{(i)})$ quantifies the degree of match between the desired and model outputs, with higher values indicating better personalization performance. All notations are summarized in ~\autoref{tab:notation}.

\begin{myremark_grey}

{{\hspace{0.05cm}\raisebox{-0.3\height+0.25ex}{\includegraphics[width=0.47cm]{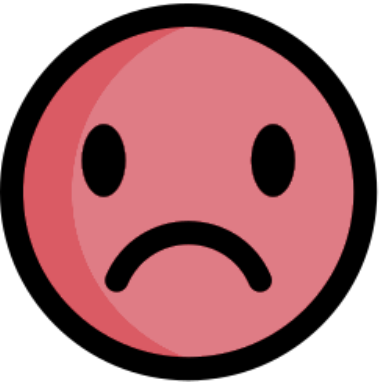}} \color{darkgray}\textbf{  One-Size-Fits-All Flaw  }}}

\vspace{-0.6em}
\hspace{2em}{\color{darkgray!60}\rule{10em}{1.25pt}}
\vspace{0.2em}

Let $M_0: \mathcal{Q} \rightarrow \mathcal{Y}$ be a base (non-personalized) LLM that maps any query $q_j^{(i)} \in \mathcal{Q}_i$ to an output $\hat{y}_{j}^{(i)} = M_0(q_j^{(i)})$. 
In the absence of personalized data $\mathcal{C}_i$, given the same query $q$, the base model $M_0$ produces identical outputs for all users. Formally:

$$\forall u_i, u_k \in \mathcal{U}, \text{ if } q \in \mathcal{Q}_i \cap \mathcal{Q}_k \text{ then } M_0(q) = y^{(i)} = y^{(k)}$$

This reflects the non-personalized nature of the base LLM, which depends only on the input query and \textbf{not on the users' personalized data, leading to large deviations from users' expectations}, i.e., large $\zeta$.

\end{myremark_grey}

\begin{myremark}

{{\hspace{0.05cm}\raisebox{-0.35\height+0.25ex}{\includegraphics[width=0.6cm]{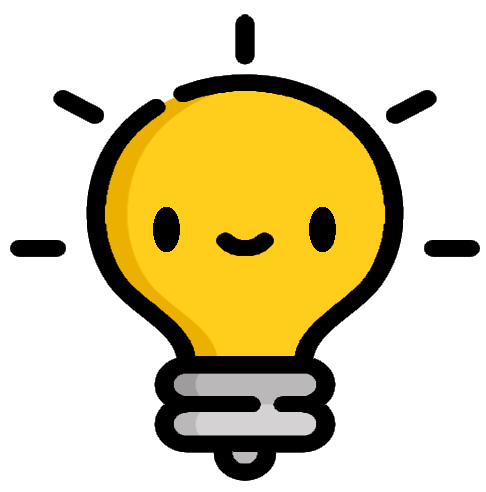}} \color{darkgreen}\textbf{Goal of PLLMs: One-Size-Fits-One  }}}

\vspace{-0.6em}
\hspace{2em}{\color{darkgreen!70}\rule{15.5em}{1.25pt}}
\vspace{0.2em}

A general framework of LLM personalization is to design the personalization operator $\mathcal{P}: (\mathcal{Q} \rightarrow \mathcal{Y}) \times \mathcal{C} \times \Theta \rightarrow (\mathcal{Q} \rightarrow \mathcal{Y})$, and optimize its parameters $\theta \subset \Theta$ (if any) to effectively inject $\mathcal{C}_i$ into the base LLM $M_0$, producing personalized models $M_i$ for user $u_i$ that minimize the overall deviation:

$$ \min_{\theta} \sum_{i=1}^{N} \sum_{j=1}^{n_i} \zeta\left( \hat{y}_j^{(i)}, y_j^{(i)} \right), $$

where, $y_j^{(i)}$ represents the personalized output generated by PLLM, $\mathcal{P}(M_0, \mathcal{C}_i; \theta)$, tailored to the user $u_i$ for query $q_j^{(i)}$. This formulation encompasses various approaches to information injection, including prompting, adaptation, and alignment methods.

\end{myremark}

Note that $\theta$ represents the tunable parameters that can exist either within the base LLM architecture itself (when fine-tuning is permitted) or within extra modules incorporated into the LLM framework, like the LoRA module~\citep{yang2024low}.  Based on personalized data $\mathcal{C}$, queries $\mathcal{Q}$, and outputs (responses) $\mathcal{Y}$, personalization can be further categorized into multiple scenarios (\autoref{fig:framework}).

\begin{figure}[htbp]
    \centering
    \includegraphics[width=0.97\linewidth]{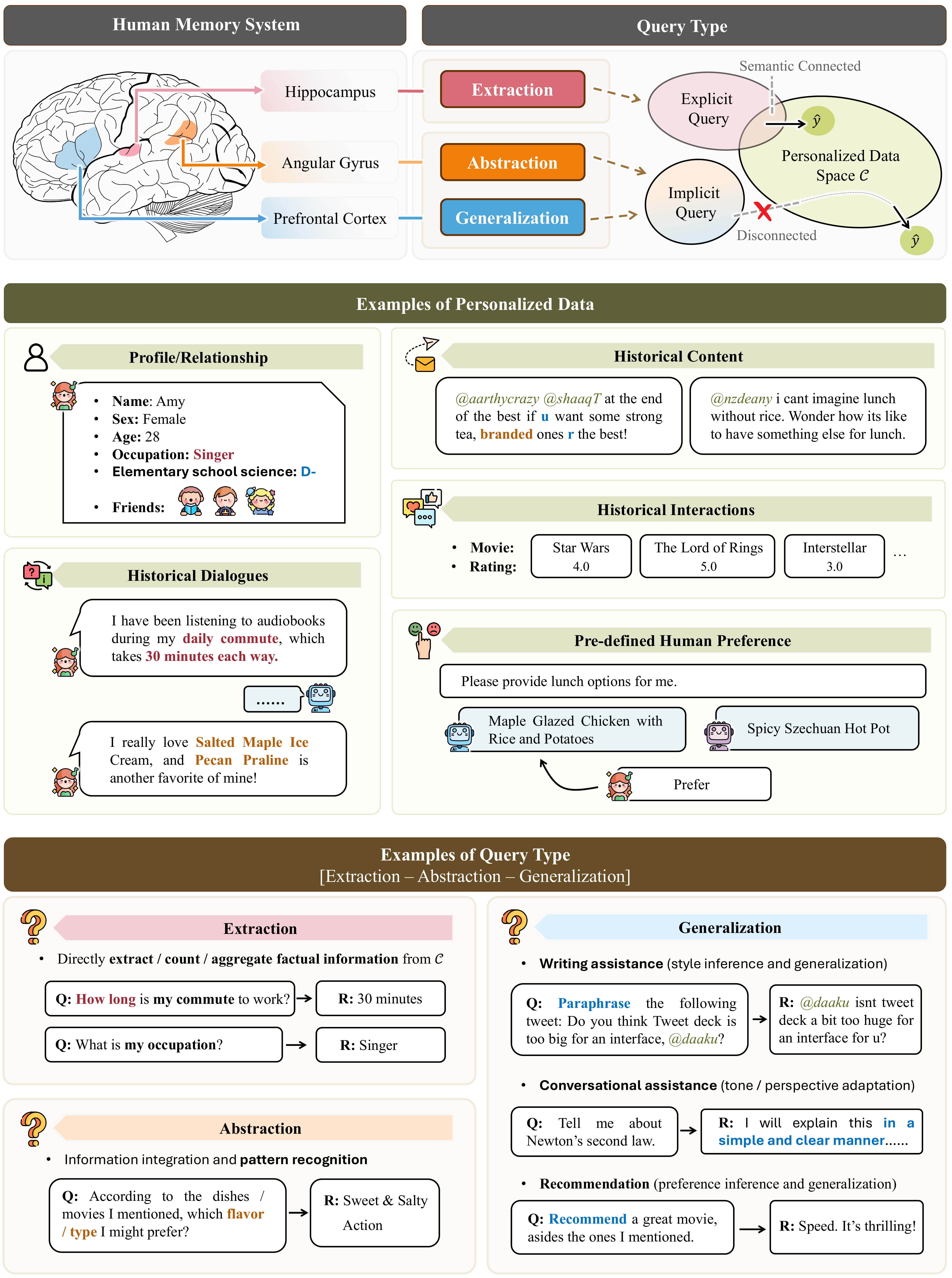}
    \caption{Examples of the personalized data and query type. The human brain regions (\textit{hippocampus} as spatiotemporal memory integrator and experience simulator, \textit{angular gyrus} as conceptual processing hub, and default mode network (including medial \textit{prefrontal cortex}) as creative ideation and remote association hub) serve as analogical references to understand these query types, though the correspondence may not be strictly one-to-one.}
    \label{fig:query type}
\end{figure}

\paragraph{Personalized Data}{\label{para:personalized data}} The personalized data $\mathcal{C}$ may encapsulate information about the user’s preferences, history, context, and other user-specific attributes. These can include profile/relationship, historical dialogues, historical content, and predefined human preferences. A detailed description, including examples, of the classification of input data types for user $u_i$ is provided:

\begin{itemize}
    \item \textbf{Profile/Relationship} User profile, including attributes (e.g., name, gender, occupation), and relationships (e.g., friends, family members), such as $\mathcal{C}_i = \{A, 18, \text{student},  \text{friends:} \{B, C, D\} \dots\}$.
    \item \textbf{Historical Dialogues} Historical dialogues, such as question-answer pairs that user $u_i$ interacts with the LLM (e.g., $\mathcal{C}_i = \{(q_0, a_0), (q_1, a_1), \dots, (q_{n_i}, a_{n_i})\}$), where each $q$ is a query and $a$ is the answer.
    \item \textbf{Historical Content} Includes documents, previous reviews, comments or feedback from user $u$. For example, $\mathcal{C}_i = \{\text{I like \textit{Avtar} because} \dots, \dots\}$.
    \item \textbf{Historical Interactions} Includes historical interactions, preferences, ratings from user $u_i$. For example, $\mathcal{C}_i = \{\text{\textit{The Lord of the Rings}}:5, \text{\textit{Interstellar}}: 3 \dots\}$.
    \item \textbf{Pre-defined Human Preference}: Define a set $S = \{d_k\}_{k=1}^K$ containing of $K$ preference dimensions such as ``Helpfulness''. Choose various combinations of these dimensions, form individual preferences, and incorporate them as the instruction. For example, a preference prompt could be ``Be harmless and helpful''.
\end{itemize}

\paragraph{Query Type} Different queries $q$ have different focal points regarding the expected responses $\hat{y}$, which have a certain influence on the method design (high-level comparison with examples is shown in ~\autoref{fig:query type} and ~\autoref{tab:query_types_comparison}).
We primarily classify them into the following categories, with further benchmark-based demonstrations provided in \autoref{sec: benchmark}:

\begin{itemize}

    \item \textbf{Extraction (explicit)}  
    This type refers to factual lookup, where the answer can be directly found in the user’s personalized data $C_i$. The model acts as a retriever:
    \begin{align*}
        \hat{y}^{(i)}_j &= f_{\mathrm{ext}}(q^{(i)}_j, C_i), \qquad
        \hat{y}^{(i)}_j \in \Sigma(C_i).
    \end{align*}
    Here $\Sigma(C_i)$ denotes the set of explicit factual tuples contained in $C_i$.  
    Importantly, extraction also covers simple statistical or aggregative operations (e.g., counting, aggregation) applied over the facts in $\Sigma(C_i)$.

    \item \textbf{Abstraction (implicit)}
    The model applies a summarization mapping $\phi: \mathcal{C} \to \mathcal{Z}$ that condenses $C_i$ into a \emph{query-independent summary} $z_i$. This abstract profile captures long-term tendencies or preferences, enabling responses beyond explicit tuples:
    \begin{align*}
        z_i &= \phi(C_i) \in \mathcal{Z}, &
        \hat{y}^{(i)}_j &= f_{\mathrm{abs}}(q^{(i)}_j, z_i), 
        & \hat{y}^{(i)}_j \notin \Sigma(C_i).
    \end{align*}

    \item \textbf{Generalization (implicit)}  
The model constructs a \emph{query-specific state} $h_{i,j}$, dynamically combining the user’s data with the current query. The response is then generated from a conditional distribution that also leverages external knowledge $\mathcal{K}$:
    \begin{align*}
        h_{i,j} &= \psi(C_i, q^{(i)}_j) \in \mathcal{H}, \\
        \hat{y}^{(i)}_j &\sim p_\theta\!\big(y \mid q^{(i)}_j, h_{i,j}, \mathcal{K}\big).
    \end{align*}
    Unlike $z_i$, which is global and query-independent, $h_{i,j}$ is dynamic and tailored to each query. External knowledge $\mathcal{K}$ is necessary when $C_i$ alone is insufficient.  
\end{itemize}

\begin{table}[t]
\small
\centering
\renewcommand{\arraystretch}{1.6}
\caption{Comparison of query types. We distinguish whether the response depends on the query itself, requires external knowledge $\mathcal{K}$, and illustrate with representative examples (\autoref{fig:query type}).}
\label{tab:query_types_comparison}
\rowcolors{2}{white}{mygrey!15}
\begin{tabular}{lccp{7cm}}   
\toprule[1.3pt]
\textbf{Type} & \textbf{Depends on query?} & \textbf{Depends on external $\mathcal{K}$?} & \textbf{Example} \\
\midrule[1.2pt]
\textbf{Extraction} & \checkmark  & \xmark  & ``What is my occupation?'' $\rightarrow$ \textit{Singer}; ``How many coffees have I bought?'' $\rightarrow$ 5 \\
\textbf{Abstraction} & \xmark  & \xmark  & ``Based on the movies I watched, what is my preference?'' $\rightarrow$ \textit{Sweet \& Salty} \\
\textbf{Generalization} & \checkmark  & \checkmark  & ``Rewrite this sentence''; ``Recommend a new movie I haven’t seen before'' \\
\bottomrule[1.3pt]
\end{tabular}
\end{table}

\paragraph{Downstream Task} By incorporating personalized data, PLLMs enhance traditional LLMs, improving response generation, recommendation, and classification tasks.

\begin{itemize}
    \item \textbf{Generation} Generation tasks typically involve $y$ representing a sequence of strings, such as generating answers for users based on their personalized data $\mathcal{C}_u$ and questions or generating content according to the user's writing style to assist their writing, and so forth~\citep{salemi2023lamp, kumar2024longlamp, zhao2025llms, au2025personalized}. 
    \item \textbf{Recommendation} The major difference between recommendation and generation is that recommendation requires suggesting specific items based on the user's historical interaction data, and it can provide reasons and explanations for the recommendations~\citep{sayana2024beyond,liu2024large}.
    \item \textbf{Classification} Classification tasks, including sentiment classification, involve labeling a particular entity (such as a movie, item, or description) based on the user's preferences to assist the user in categorization or summarization~\citep{salemi2023lamp, au2025personalized, zhao2025llms}. 
\end{itemize}

\textbf{Note that} our survey differs significantly from role-play related LLM personalization~\citep{tseng2024two, chen2024large, zhang2024personalization}. While role-play focuses on mimicking characters during conversations~\citep{zhu2024player, zhao2024narrativeplay}, PLLMs in this survey focus on understanding users' contexts and preferences to meet their specific needs. 
Compared to~\citep{zhang2024personalization}, which emphasizes broad categories, our work provides a quite different systematic analysis of techniques to enhance PLLM efficiency and performance, with a detailed technical classification.

\subsection{Proposed Taxonomy}
\label{sec:taxonomy}

We propose a taxonomy (as illustrated in \autoref{fig:framework} and \autoref{fig:taxonomy_of_PLLM}) from technical perspectives (i.e., regarding the personalization operation $\mathcal{P}$), categorizing the methods for Personalized Large Language Models (PLLMs) into three major levels: 

\begin{enumerate}[label={(\arabic*)}]
    \item \textbf{Input level: Personalized Prompting} focuses on handling user-specific data outside the LLM and injecting it into the model (\autoref{sec: personalized prompting}).
    \item \textbf{Model level: Personalized Adaptation} emphasizes designing a framework to efficiently fine-tune or adapt model parameters for personalization (\autoref{sec: personalized adatation}).
    \item \textbf{Objective Level: Personalized Alignment} aims to refine model behavior to align with user preferences effectively (\autoref{sec: personalized alignment}).
\end{enumerate}

\section{Personalized Prompting}
\label{sec: personalized prompting}

\begin{figure*}[!t]
    \centering
    \includegraphics[width=0.96\linewidth]{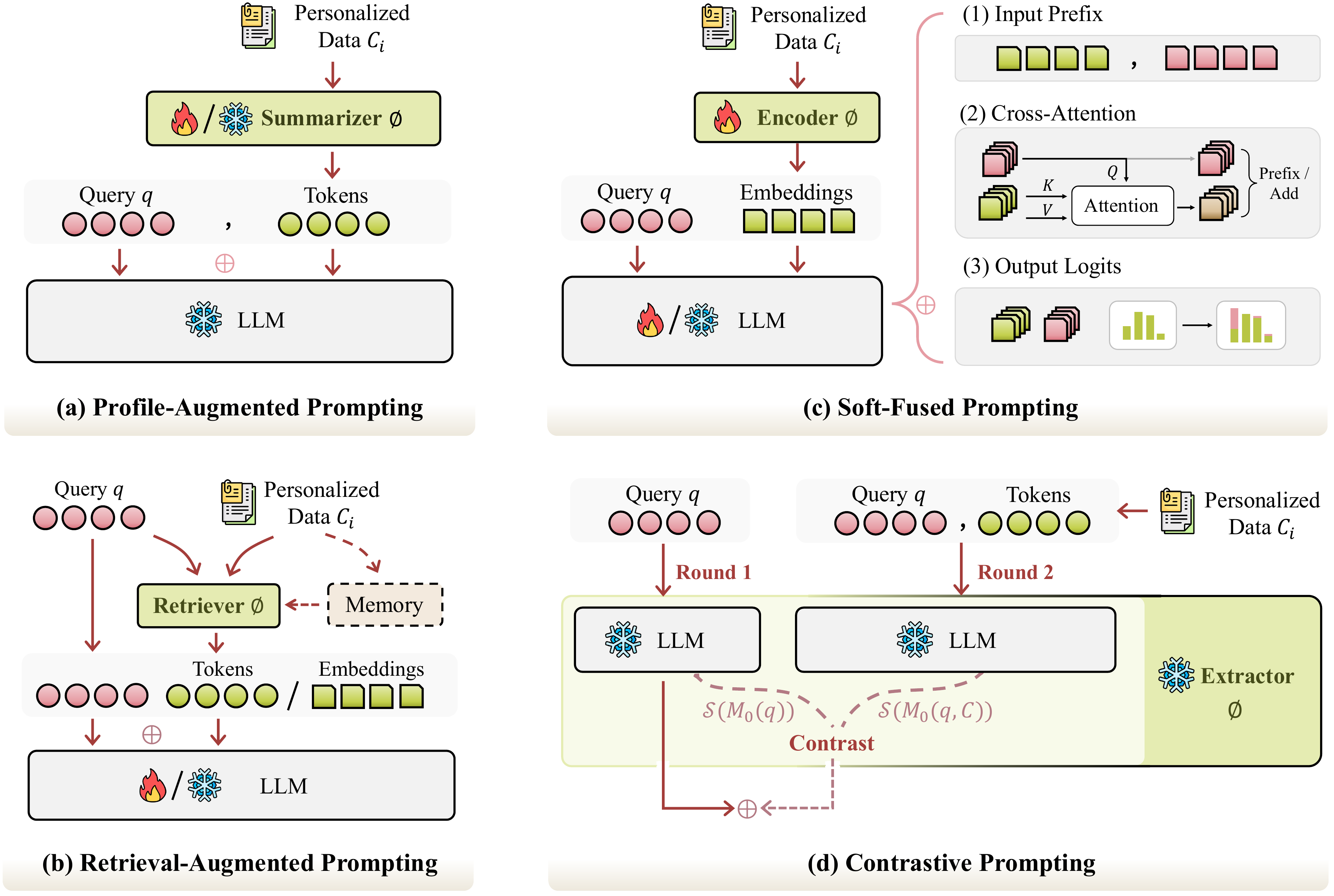}
    \caption{
    The illustration of personalized prompting approaches: \textbf{(a) Profile-Augmented Prompting} summarizes personalized data into tokens and concatenates with queries; \textbf{(b) Retrieval-Augmented Prompting} retrieves relevant records from memory and combines with queries; \textbf{(c) Soft-Fused Prompting} encodes personalized data into embeddings integrated via input prefix, cross-attention, or output logits; \textbf{(d) Contrastive Prompting} compares model outputs with and without personalized information to extract personalization factors.
    }
    \label{fig:prompting}
\end{figure*}

Prompt engineering acts as a bridge for interaction between users and LLMs. In this survey, prompting involves guiding an LLM to generate desired outputs using various techniques, from traditional text prompts to advanced methods like soft embedding. Soft embedding can be extended not only through input but also via cross-attention or by adjusting output logits, enabling more flexible and context-sensitive responses.
For each user $u_i$, the personalization operation $\mathcal{P}$ is expressed as
\begin{equation}
y_j^{(i)} = M_0 \left(q_j^{(i)} \oplus \phi \left( \mathcal{C}_i\right); \theta \right)
\label{equ:prompting}
\end{equation}
where, 
$\phi$ is a function that extracts relevant context from the user's personal context $\mathcal{C}_i$;
$\oplus$ represents the combination operator that fuses the query $q_{j}^{(i)}$ and the relevant personalized context $\phi \left(\mathcal{C}_i \right)$, producing enriched information for the LLM. Based on different designs of the $\phi$ operation, we further categorize prompting-based methods into four types, as shown in ~\autoref{fig:prompting} and ~\autoref{tab: prompting operation}: (1) profile-augmented prompting; (2) retrieval-augmented prompting; (3) soft-fused prompting; and (4) contrastive prompting.

\subsection{Profile-Augmented Prompting}
\label{subsec:PAG}

Profile-augmented prompting (\autoref{fig:prompting}  (a)) explicitly utilizes summarized user preferences and profiles in natural language to augment LLMs’ input at the token level ($\phi$ is the \textit{summarizer} model). The summarizer model is typically distinct from the base LLM (generator).

\textbf{Non-tuned Summarizer}\quad A frozen LLM can be directly used as the summarizer to summarize user profiles due to its strong language understanding capabilities, i.e., $\mathcal{\phi}\left(\mathcal{C}_u\right) = M \left(\mathcal{C}_u\right)$. 
For instance, \textbit{Cue-CoT}~\citep{wang2023cue} employs chain-of-thought prompting for personalized profile augmentation, using LLMs to extract and summarize user status (e.g., emotion, personality, and psychology) from historical dialogues. 
\textbit{PAG}~\citep{richardson2023integrating} leverages instruction-tuned LLMs to pre-summarize user profiles based on historical content. The summaries are stored offline, enabling efficient personalized response generation while meeting runtime constraints.
\textbit{ONCE}~\citep{DBLP:conf/wsdm/LiuCS024} prompts closed-source LLMs to summarize topics and regions of interest from users' browsing history, enhancing personalized recommendations.
\textbit{DPL}~\citep{qiu2025measuring} enhances personalization through inter-user comparison, identifying similar users in the same cluster. Leveraging central users from the same cluster where the target user is located creates more accurate preference summaries.

\textbf{Tuned Summarizer}\quad 
Black-box LLMs are sensitive to input noise, like off-topic summaries, and struggle to extract relevant information. Thus, training the summarizer to adapt to user preferences and style is essential.
\textbit{Matryoshka}~\citep{li2024matryoshka} uses a white-box LLM to summarize user histories, similar to PAG, but fine-tunes the summarizer instead of the generator LLM.
\textbit{RewriterSlRl}~\citep{li2024learning}  rewrites the query $q$ instead of concatenating summaries, optimized with supervised and reinforcement learning.

\begin{table}[!t]
    \centering
    \renewcommand\arraystretch{1.5}
    \small
    \caption{Summarization of the operation difference among different prompting methods in \autoref{equ:prompting}. $\mathcal{S}$ denotes the operation that extracts inner information from LLM. Black-box refers to whether the method can be applied to black-box models, where the base model parameters are not accessible.  }
    \vspace{4pt}
    \resizebox{\textwidth}{!}{
        \begin{tabular}{lcccc}
        \toprule[1.3pt]
        & \parbox[c]{4cm}{\centering \textbf{Profile-Augmented} \\ (\autoref{subsec:PAG}) } & \parbox[c]{4cm}{\centering \textbf{Retrieval-Augmented} \\ (\autoref{subsec:RAG}) }  & \parbox[c]{4cm}{\centering \textbf{Soft-Fused} \\ (\autoref{subsec:soft}) } & \parbox[c]{4cm}{\centering \textbf{Contrastive} \\ (\autoref{subsec: contrastive prompting}) } \\
        \midrule[1.3pt]
        $\phi$ & Summerizer ($M$) & Retriever & Encoder &  \parbox[c]{4cm}{\centering Extractor \\ $\mathcal{S}\left((M_0(q, \mathcal{C}\right)) , \mathcal{S}\left(M_0(q)\right)$ }\\
        \arrayrulecolor{gray} \midrule[0.6pt]
        $\oplus$ & Token Combination & Token Combination & Input Prefix; Attention; Logits & Hidden States; Logits \\
        \midrule[1.0pt]
        
        \rowcolor{mygreen2!35}
        \textbf{Pros.} & {\color{darkgreen}\textbf{Efficiency}} & {\color{darkgreen}\textbf{Long-Term Memory}} & {\color{darkgreen}\textbf{Semantic Nuances}} & \parbox[c]{4cm}{\centering {\color{darkgreen}\textbf{Interpretability} \\ \textbf{Controllable}}} \\
        \arrayrulecolor{gray}\midrule[0.6pt]
        
        \rowcolor{mygrey!20}
        \textbf{Cons.} & {\color{darkgray!90}\textbf{Infromation Loss}} & \parbox[c]{4cm}{\centering {\color{darkgray!90}\textbf{Computational Limits}} \\ {\color{darkgray!90}\textbf{Irrelevant Data}} } & {\color{darkgray!90}\textbf{Lack of Interpretability}} & {\color{darkgray!90}\textbf{Hyperparameter Sensitivity}} \\
        \bottomrule[1.3pt]
        \end{tabular}
    }
    \label{tab: prompting operation}
\end{table}

\subsection{Retrieval-Augmented Prompting}
\label{subsec:RAG}

Retrieval-augmented prompting~\citep{gao2023retrieval, fan2024survey,qiu2024entropy} excels at extracting the most relevant records from user data to enhance PLLMs (See \autoref{fig:prompting} (b)).
Due to the complexity and volume of user data, many methods use an additional \textit{memory} for more effective retrieval~\citep{tan2025membench}. Common retrievers include sparse (e.g., BM25~\citep{robertson1995okapi}),  and dense retrievers (e.g., Faiss~\citep{johnson2019billion}, Contriever~\citep{izacard2021unsupervised}).
These methods effectively manage the increasing volume of user data within the LLM's context limit, improving relevance and personalization by integrating key evidence from the user's personalized data.

\subsubsection{Personalized Memory Construction} 

This part designs mechanisms for retaining and updating memory to enable efficient retrieval of relevant information.

\textbf{Non-Parametric Memory}\quad 
This category maintains a token-based database, storing and retrieving information in its original tokenized form without using parameterized vector representations. For example, 
\textbit{MemPrompt}~\citep{madaan2022memory} and \textbit{TeachMe}~\citep{dalvi2022towards} maintain a dictionary-based feedback memory (key-value pairs of mistakes and user feedback).
MemPrompt focuses on prompt-based improvements, whereas TeachMe emphasizes continual learning via dynamic memory that adapts over time.
\textbit{MaLP}~\citep{zhang2024llm} further integrates multiple memory types, leveraging working memory for immediate processing, short-term memory (STM) for quick access, and long-term memory (LTM) for key knowledge. \textbit{FERMI}~\citep{kim2024few} maintains misaligned responses and creates personalized prompts for users by using LLMs to progressively refine prompting strategies based on user profiles and past opinions.

\textbf{Parametric Memory}\quad
Recent studies parameterize and project personalized user data into a learnable space, with parametric memory filtering out redundant context to reduce noise.
For instance, 
\textbit{LD-Agent}~\citep{li2024hello} maintains memory with separate short-term and long-term banks, encoding long-term events as parametric vector representations refined by a tunable module and retrieved via an embedding-based mechanism.
\textbit{MemoRAG}~\citep{qian2024memorag}, in contrast, adopts a different approach by utilizing a lightweight LLM as memory to learn user-personalized data. Instead of maintaining a vector database for retrieval, it generates a series of tokens as a draft to further guide the retriever, offering a more dynamic and flexible method for retrieval augmentation. Inspired by cognitive theories of memory, \textit{PRIME}~\citep{zhang2025prime} introduces episodic and semantic memory mechanisms to enhance the personalization capabilities of LLMs, and employs a slow-thinking strategy to further improve personalized reasoning.

\subsubsection{Personalized Memory Retrieval}

The key challenge in the personalized retriever design lies in selecting not only relevant but also representative personalized data for downstream tasks. \textbit{LaMP}~\citep{salemi2023lamp} investigates how retrieved personalized information affects the responses of large language models (LLMs) through two mechanisms: in-prompt augmentation (IPA) and fusion-in-decoder (FiD).
\textbit{PEARL}~\citep{mysore2023pearl} and \textbit{ROPG}~\citep{salemi2024optimization} similarly aim to enhance the retriever using personalized generation-calibrated metrics, improving both the personalization and text quality of retrieved documents.
Meanwhile, \textbit{HYDRA}~\citep{zhuang2024hydra} trains a reranker to prioritize the most relevant information additionally from top-retrieved historical records for
enhanced personalization.
\textbit{RPM}~\citep{kim2025llms} focuses on reasoning ability by extracting user-specific factors from user historical personalized data and creating annotated reasoning paths. At inference, it retrieves these examples to guide reasoning-aligned outputs, strengthening personalized responses.
\textbit{HYDRA} and \textbit{RPM} primarily target black-box models where access to the base LLM parameters is restricted.

\subsection{Soft-Fused Prompting}
\label{subsec:soft}

Soft prompting differs from profile-augmented prompting in that it compresses personalized data into soft embeddings rather than summarizing it into discrete tokens.
These embeddings
are generated by a user feature \textit{encoder} $\phi$. 

In this survey, we generalize the concept of soft prompting, showing that soft embeddings can be integrated (combination operator $\oplus$) not only through the input but also via cross-attention or by adjusting output logits, allowing for more flexible and context-sensitive responses (See \autoref{fig:prompting} (c)).

\textbf{Input Prefix}\quad
Soft prompting, used as an input prefix, focuses on the embedding level by concatenating the query embedding with the soft embedding, and is commonly applied in recommendation tasks.
\textbit{PPlug}~\citep{liu2024llms+} constructs a user-specific embedding for each individual by modeling their historical contexts using a lightweight plug-in user embedder module. This embedding is then attached to the task input.
\textbit{UEM}~\citep{doddapaneni2024user} is a user embedding module (transformer network) that generates a soft prompt conditioned on the user's personalized data. 
\textbit{PERSOMA}~\citep{hebert2024persoma} enhances UEM by employing resampling, selectively choosing a subset of user interactions based on relevance and importance. 
\textbit{REGEN}~\citep{sayana2024beyond} combines item embeddings from user-item interactions via collaborative filtering and item descriptions using a soft prompt adapter to generate contextually personalized responses. 
\textbit{PeaPOD}~\citep{ramos2024preference} personalizes soft prompts by distilling user preferences into a limited set of learnable, dynamically weighted prompts. Unlike previously mentioned methods, which focus on directly embedding user interactions or resampling relevant data, PeaPOD adapts to user interests by weighting a shared set of prompts. 
\textbit{ComMer}~\citep{zeldes2025commer} tunes an encoder to compress historical documents into embeddings, merges them through mean pooling, then feeds the result into a frozen base LLM, distinguishing it from methods focused on historical interaction data. \textbit{DEP} ~\citep{qiu2025latent}  further models inter-user differences in the latent space instead of merely considering the user's own information.

\textbf{Cross-Attention}\quad
Cross-attention enables the model to process and integrate multiple input sources by allowing it to attend to personalized data and the query. 
\textbit{User-LLM}~\citep{ning2024user} uses an autoregressive user encoder to convert historical interactions into embeddings through self-supervised learning, which are then integrated via cross-attention. The system employs joint training to optimize both the retriever and generator for better performance.
\textbit{RECAP}~\citep{liu2023recap} utilizes a hierarchical transformer retriever designed for dialogue domains to fetch personalized information. This information is integrated into response generation via a context-aware prefix encoder, improving the model's ability to generate personalized, contextually relevant responses.

\textbf{Output Logits}\quad
\textbit{GSMN}~\citep{wu2021personalized} retrieves relevant information from personalized data, encodes it into soft embeddings, and uses them in attention with the query vector. Afterward, the resulting embeddings are concatenated with the LLM-generated embeddings, modifying the final logits to produce more personalized and contextually relevant responses.

\subsection{Contrastive Prompting}
\label{subsec: contrastive prompting}

The key insight of contrastive prompting is to utilize two forward paths of LLM to generate contrast pairs with and without personalized information (See~\autoref{fig:prompting} (d)). By comparing model states $S$ with and without personalization, it identifies factors influencing personalization, enabling dynamic adjustment of the model's personalization level. Two mainstream model states are utilized: \textbf{hidden states} (representations) and \textbf{logits} (predictive distributions). 

\textbit{CoS}~\citep{he2024cos} is a special case that assumes there is a brief user profile $\mathcal{C}$ for each query; it
amplifies its influence in LLM response generation by combining output probabilities (logits) with and without the profile, i.e., $\mathcal{S}\left(M_0(q)\right) + \alpha \times \mathcal{S}\left(M_0(q, \mathcal{C})\right)$, adjusting personalization degree through hyperparameter $\alpha$ without fine-tuning. 
In contrast, \textbit {StyleVector}~\citep{zhang2025personalized} obtains encoded personalized information in LLM via hidden representations. Specifically, given a user $u_i$ and the personalized data (user historical content), $\left(c_j^{(i)}, r_j^{(i)}\right) \in \mathcal{C}_i$, it use GPT-3.5-Turbo $M$ to generate a general response for comparison with the expected personalized response $r_j^{(i)}$, i.e., $\phi: \frac{1}{|\mathcal{C}_i|}\sum_{j=1}^{|\mathcal{C}_i|} \left[\mathcal{S}\left( M_0\left(c_j^{(i)}, M\left(c_j^{(i)}\right)\right)\right) - \mathcal{S}\left(M_0 \left( c_j^{(i)}, r_j^{(i)} \right) \right) \right]$, where $\mathcal{S}$ extracts hidden representations in base model $M_0$'s middle layer. $\phi$ extracts the user style-based vector, and by adding this vector with a scaling factor $\alpha$, this method can steer the LLM toward personalized output to a controllable degree. \textbit{CoSteer}~\citep{lv2025costeer} performs decoding-time personalization through \emph{local delta steering}. A local SLM produces logits with and without personal context, and their difference $\Delta=\log \pi^*_{\text{pers}}-\log \pi^*_{\text{base}}$ is used to steer the cloud LLM’s logits iteratively, preserving privacy while achieving controllable personalization without fine-tuning.

\subsection{Discussions}\quad
The three prompting methods have distinct pros and cons (\autoref{tab: prompting operation}): 

\begin{myinfobox}

{{\hspace{0.05cm}\raisebox{-0.35\height+0.25ex}{\includegraphics[width=0.6cm]{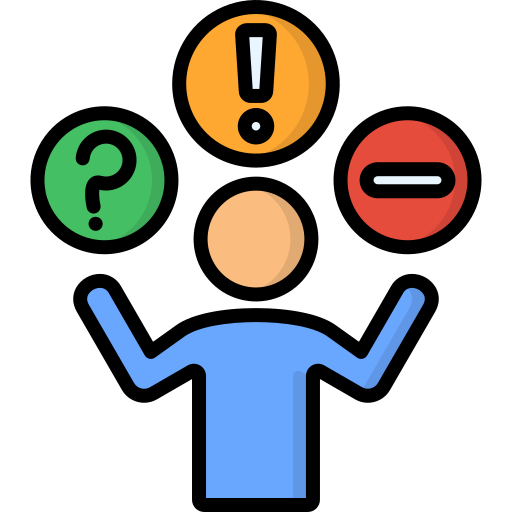}} \color{darkgreen}\textbf{Personalized Prompting Method: Pros and Cons}}}

\begin{itemize}
    \item Profile-augmented prompting improves efficiency by compressing historical data but risks information loss and reduced personalization.
    \item Retrieval-augmented prompting offers rich, context-aware inputs and scales well for long-term memory but can suffer from computational limits and irrelevant data retrieval. 
    \item Soft prompting efficiently embeds user-specific info, capturing semantic nuances without redundancy, but is limited to black-box models and lacks explicit user preference analysis.
    \item Contrastive prompting provides interpretability and controllable personalization by comparing model states with and without personalized information, but suffers from hyperparameter sensitivity when choosing the hyperparameter, i.e., scaling factor $\alpha$. 
\end{itemize}

\end{myinfobox}

This approach centers on maintaining, selecting, and integrating external personalized data into LLMs to improve user behavior comprehension. Overall, prompting-based methods are efficient and adaptable, enabling dynamic personalization with minimal computational overhead. Especially for RAG, which reduces LLM hallucination by grounding responses in retrieved evidence. Therefore, prompting methods are particularly suitable for \textit{explicit queries} that require answering specific factual information, as mentioned in \autoref{sec: problem statement}, excelling when personalization involves retrieving concrete information from user data.

However, they lack deeper personalization analysis, as they rely on predefined prompt structures to inject user-specific information and are limited in accessing global knowledge due to the narrow scope of prompts, which usually fail in tasks with \textit{implicit generation queries} compared with adaptation and alignment methods~\citep{DBLP:conf/emnlp/Tan000Y024}.

\section{Personalized Adaptation}
\label{sec: personalized adatation}

\begin{figure*}[!t]
    \centering
    \includegraphics[width=0.99999\linewidth]{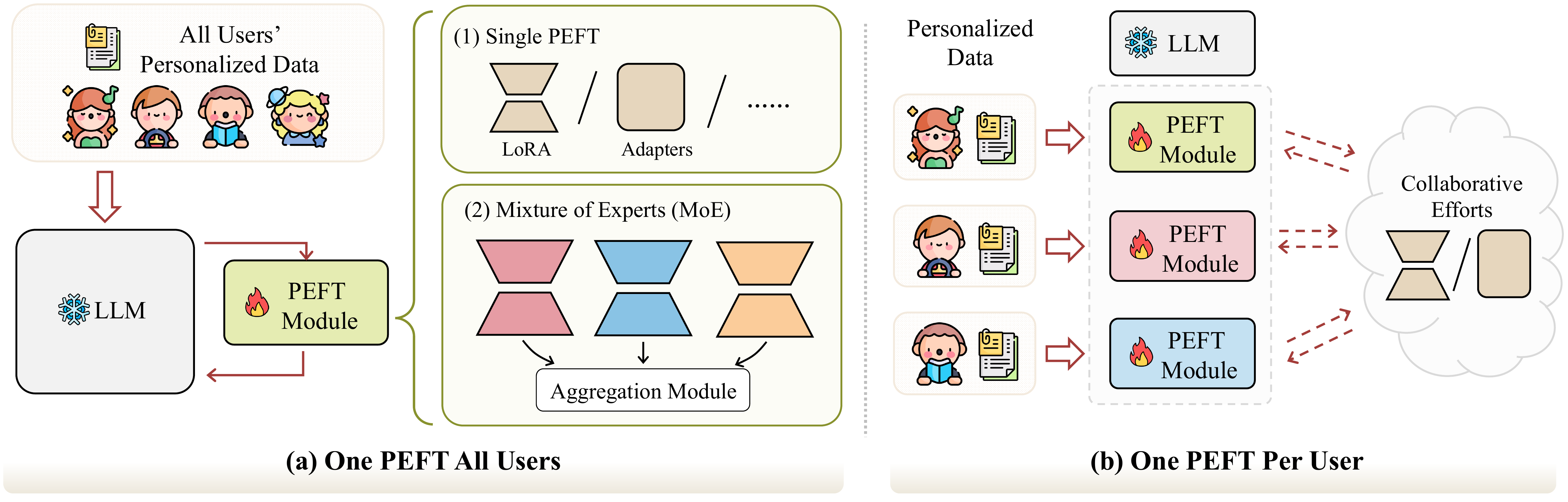}
    \caption{Illustration of personalized adaptation approaches: \textbf{(a) One PEFT for All Users} with shared parameters (\autoref{subsec:one4all}), and \textbf{(b) One PEFT Per User} with individual parameters and optional collaborative efforts (\autoref{subsec:one4one}).}
    \label{fig:peft}
\end{figure*}

\begin{table}[!t]
    \centering
    \renewcommand\arraystretch{1.5}
    \small
    \caption{Comparison of two PEFT approaches for PLLMs. The training objectives and inference mechanisms differ fundamentally in parameter sharing strategies.}
    \vspace{4pt}
    \resizebox{\textwidth}{!}{
        \begin{tabular}{lcc}
        \toprule[1.3pt]
    & \parbox[c]{6cm}{\centering \textbf{One PEFT for All Users} \\ (Shared Parameters,~\autoref{subsec:one4all}) } & \parbox[c]{6cm}{\centering \textbf{One PEFT Per User} \\ (Individual Parameters,~\autoref{subsec:one4one}) } \\
        \midrule[1.3pt]
        \textbf{Training Objective} & $\displaystyle\min_{\theta} \sum_{i=1}^N \sum_{j=1}^{|\mathcal{C}_i|} \mathcal{L}\big(M_0(c^{(i)}_j;\theta), r^{(i)}_j \big)$ & $\displaystyle\min_{\{\theta_i\}_{i=1}^N} \sum_{i=1}^N \sum_{j=1}^{|\mathcal{C}_i|} \mathcal{L}\big(M_0(c^{(i)}_j;\theta_i), r^{(i)}_j \big)$ \\
        \arrayrulecolor{gray} \midrule[0.6pt]
        \textbf{Inference Stage} & $y_j^{(i)} = M_0(q^{(i)}_j, \mathcal{C}_i; \theta)$ & $y_j^{(i)} = M_0(q^{(i)}_j; \theta_i)$ \\
        \midrule[1.0pt]
        
        \rowcolor{mygreen2!35}
        \textbf{Pros.} & \parbox[c]{6cm}{\centering {\color{darkgreen}\textbf{Parameter Efficiency}} \\ {\color{darkgreen}\textbf{Scalability}}} & \parbox[c]{6cm}{\centering {\color{darkgreen}\textbf{Strong Personalization}} \\ {\color{darkgreen}\textbf{User Isolation}}} \\
        \arrayrulecolor{gray}\midrule[0.6pt]
        
        \rowcolor{mygrey!20}
        \textbf{Cons.} & \parbox[c]{6cm}{\centering {\color{darkgray!90}\textbf{Limited Personalization}} \\ {\color{darkgray!90}\textbf{Persnonalized Data Dependency}}} & \parbox[c]{6cm}{\centering {\color{darkgray!90}\textbf{Storage Overhead}} \\ {\color{darkgray!90}\textbf{Training Complexity}}} \\
        \bottomrule[1.3pt]
        \end{tabular}
    }
    \label{tab:peft_comparison}
\end{table}

PLLMs require balancing fine-tuning's deep adaptability with the efficiency of prompting.
Therefore, specialized methods need to be specifically designed for PLLMs to address these challenges utilizing parameter-efficient fine-tuning methods (PEFT),  such as LoRA~\citep{hu2021lora, yang2024low},
prefix-tuning~\citep{li2021prefix}, MeZo~\citep{malladi2023fine}, etc.  From an architectural perspective, these methods can be categorized into two types: (1) one PEFT for all users, where all users share a single fine-tuning module that differentiates personalized information, and (2) one PEFT per user, where each user has their own unique fine-tuning block. Details are illustrated in \autoref{fig:peft} and ~\autoref{tab:peft_comparison}.

\subsection{One PEFT All Users}
\label{subsec:one4all}

This method trains on all users' data using a \textit{shared PEFT module}, eliminating the need for separate modules per user. The shared module's architecture can be further categorized.

\textbf{Single PEFT}\quad
\textbit{PLoRA}~\citep{DBLP:conf/aaai/ZhangWYXZ24} and \textbit{LM-P}~\citep{wozniak2024personalized} 
utilize LoRA for PEFT of LLM, injecting personalized information via user embeddings and user IDs, respectively. PLoRA is further extended and supports online training and prediction for cold-start scenarios. \textbit{UserIdentifier}~\citep{mireshghallah2021useridentifier} uses a static, non-trainable user identifier to condition the model on user-specific information, avoiding the need for trainable user-specific parameters and reducing training costs. 
\textbit{Review-LLM}~\citep{peng2024llm} aggregates users' historical behaviors and ratings into prompts to guide sentiment and leverages LoRA for efficient fine-tuning.
However, these methods rely on a single architecture with fixed configurations (e.g., hidden size, insertion layers), making them unable to store and activate diverse information for personalization~\citep{zhou2024autopeft}. To solve this problem,
\textbit{MiLP}~\citep{zhang2024personalized} utilizes a Bayesian optimization strategy to automatically identify the optimal configuration for applying multiple LoRA modules, enabling efficient and flexible personalization.

\textbf{Mixture of Experts (MoE)}\quad
Several methods use the LoRA module, but with a static configuration for all users. This lack of parameter personalization limits adaptability to user dynamics and preference shifts, potentially resulting in suboptimal performance~\citep{cai2024survey}. \textbit{RecLoRA}~\citep{zhu2024lifelong} addresses this limitation by maintaining a set of parallel, independent LoRA weights and employing a soft routing method to aggregate meta-LoRA weights, enabling more personalized and adaptive results. Similarly, 
\textbit{iLoRA}~\citep{kongcustomizing} creates a diverse set of experts (LoRA) to capture specific aspects of user preferences and generates dynamic expert participation weights to adapt to user-specific behaviors.

Shared PEFT methods rely on a centralized approach, where user-specific data is encoded into a shared adapter by centralized LLMs. This limits the model's ability to provide deeply personalized experiences tailored to individual users. Furthermore, using a centralized model often requires users to share personal data with service providers, raising concerns about the storage, usage, and protection of this data.

\vspace{-0.5cm}
\subsection{One PEFT Per User}
\label{subsec:one4one}
\vspace{-0.5cm}

Equipping \textit{a user-specific PEFT module} makes LLM deployment more personalized while preserving data privacy. However, the challenge lies in ensuring efficient operation in resource-limited environments, as users may lack sufficient local resources to perform fine tuning. 

\textbf{No Collaboration}\quad 
There is no collaboration or coordination between adapters or during the learning process for each use in this category.
\textbit{UserAdapter}~\citep{zhong2021useradapter} personalizes models through prefix-tuning, fine-tuning a unique prefix vector for each user while keeping the underlying transformer model shared and frozen. 
\textbit{PocketLLM}~\citep{peng2024pocketllm} utilizes a derivative-free optimization approach, based on MeZo~\citep{malladi2023fine}, to fine-tune LLMs on memory-constrained mobile devices. 

\textbf{Collaborative Efforts}\quad
The ``one-PEFT-per-user" paradigm without collaboration is computationally and storage-intensive, particularly for large user bases. Additionally, individually owned PEFTs hinder community value, as personal models cannot easily share knowledge or benefit from collaborative improvements.
\textbit{OPPU}~\citep{DBLP:conf/emnlp/Tan000Y024} frames the personalized LLMs training as a two-stage process, where a global LoRA module is learned to capture shared knowledge across users initially, followed by user-specific LoRA modules for individual adaptation. PerFit~\citep{liu2025exploring} further enhances this by replacing LoRA modules with representation fine-tuning modules, motivated by their finding that the personalization of LLM correlates with a low-rank collaborative shift and diverse personalized shifts following the collaborative one.~\textbit{CoPe}~\citep{bu2025personalized} utilizes reward-guided decoding tailored for personalization, aiming to maximize the implicit reward signal for each user.
\textbit{PER-PCS}~\citep{tan2024personalized} enables efficient and collaborative PLLMs by sharing a small fraction of PEFT parameters across users. It first divides PEFT parameters into reusable pieces with routing gates and stores them in a shared pool. For each target user, pieces are autoregressively selected from other users, ensuring scalability, efficiency, and personalized adaptation without additional training. \textbit{PROPER}~\citep{zhang2025proper}  introduces hierarchical user grouping with MoE-LoRA integration, enabling fine-grained personalization through staged adaptation from population-level, group-level, to individual-level models.

Another efficient collaborative strategy is based on the federated learning (FL) framework.
For example, 
\cite{wagner2024personalized} introduces a FL framework for on-device LLM fine-tuning, using strategies to aggregate LoRA model parameters and handle data heterogeneity efficiently, outperforming purely local fine-tuning. 
\textbit{FDLoRA}~\citep{qi2024fdlora} introduces a personalized FL framework using dual LoRA modules to capture personalized and global knowledge. It shares only global LoRA parameters with a central server and combines them via adaptive fusion, enhancing performance while minimizing communication and computing costs.

There are other frameworks that can be explored, such as \textbit{HYDRA}~\citep{zhuang2024hydra}, which also employs a base model to learn shared knowledge. However, in contrast to federated learning, it assigns distinct heads to each individual user to extract personalized information.

\subsection{Discussions}

Fine-tuning methods enable deep personalization by modifying a large set of model parameters, and parameter-efficient fine-tuning methods (e.g., prefix vectors or adapters) reduce computational cost and memory requirements while maintaining high personalization levels. These methods improve task adaptation by tailoring models to specific user needs, enhancing performance in tasks like sentiment analysis and recommendations. They also offer flexibility, allowing user-specific adjustments while leveraging pre-trained knowledge. 
However, they still face several challenges:

\begin{myinfobox}

{{\hspace{0.05cm}\raisebox{-0.35\height+0.25ex}{\includegraphics[width=0.6cm]{Figures/icons/issue.png}} \color{darkgreen}\textbf{Personalized Adaptation Methods: Challenges}}}

    \paragraph{Overfitting} With limited or noisy user data, models may fail to generalize and lose robustness.
    \paragraph{Performance–Privacy–Efficiency Trade-off} A per-user PEFT strategy ensures strong privacy but suffers from limited performance. Introducing collaboration improves performance yet risks privacy leakage. Sharing only parameters instead of raw user data alleviates privacy concerns, but introduces efficiency challenges in edge–cloud coordination, underscoring the difficulty of jointly optimizing all three dimensions.
    \paragraph{Cold-start} Adapting to new users with sparse or unseen data remains difficult.

\end{myinfobox}

Addressing these issues calls for complementary strategies such as federated learning, synthetic data generation, and hybrid prompting--adaptation pipelines, in order to balance efficiency, personalization depth, and reliability.

\section{Personalized Alignment}
\label{sec: personalized alignment}

Alignment techniques~\citep{bai2022training,rafailov2024direct} typically optimize LLMs to match the generic preferences of humans. However, in reality, individuals may exhibit significant variations in their preferences for LLM responses across different dimensions like language style, knowledge depth, and values. Personalized alignment seeks to further align with individual users’ unique preferences beyond generic preferences. 
A significant challenge in personalized alignment is creating high-quality user-specific preference datasets, which are more complex than general alignment datasets due to data sparsity. The second challenge arises from the need to refine the canonical RLHF framework~\citep{ouyang2022training} to handle the diversification of user preferences, which is essential for integrating personalized preferences without compromising efficiency and performance.

\subsection{Data Construction}
\label{subsec:data}

High-quality data construction is critical for learning PLLMs, primarily involving self-generated data through interactions with the LLM. \citeauthor{wu2024aligning}~\citep{wu2024aligning} constructs a dataset for aligning LLMs with individual preferences by initially creating a diverse pool of 3,310 user personas, which are expanded through iterative self-generation and filtering.  This method is similar to \textbit{PLUM}~\citep{magister2024way} 
that both simulate dynamic interactions through multi-turn conversation trees, allowing LLMs to infer and adapt to user preferences. To enable LLMs to adapt to individual user preferences without re-training, \citeauthor{lee2024aligning}~\citep{lee2024aligning} utilizes diverse system messages as meta-instructions to guide the models' behavior. To support this, the MULTIFACETED COLLECTION dataset is created, comprising 197k system messages that represent a wide range of user values. To facilitate real-time, privacy-preserving personalization on edge devices while addressing data privacy, limited storage, and minimal user disruption, \citeauthor{qin2024enabling}~\citep{qin2024enabling} introduces a self-supervised method that efficiently selects and synthesizes essential user data, improving model adaptation with minimal user interaction.

Research efforts are also increasingly concentrating on developing datasets that assess models' comprehension of personalized preferences. \citeauthor{kirk2024prism}~\citep{kirk2024prism} introduces \textbit{PRISM Alignment Dataset} that maps the sociodemographics and preferences of 1,500 participants from 75 countries to their feedback in live interactions with 21 LLMs, focusing on subjective and multicultural perspectives on controversial topics. \textbit{PersonalLLM}~\citep{zollo2024personalllm} introduces a novel personalized testdb, which curates open-ended prompts and multiple high-quality responses to simulate diverse latent preferences among users. It generates simulated user bases with varied preferences from pre-trained reward models, addressing the challenge of data sparsity in personalization. \textbit{ALOE}~\citep{wu2024aligning} further generates tree-structured multi-turn conversations instead of  single-turn pairwise responses.

\begin{figure*}[!t]
    \centering
    \includegraphics[width=0.99999\linewidth]{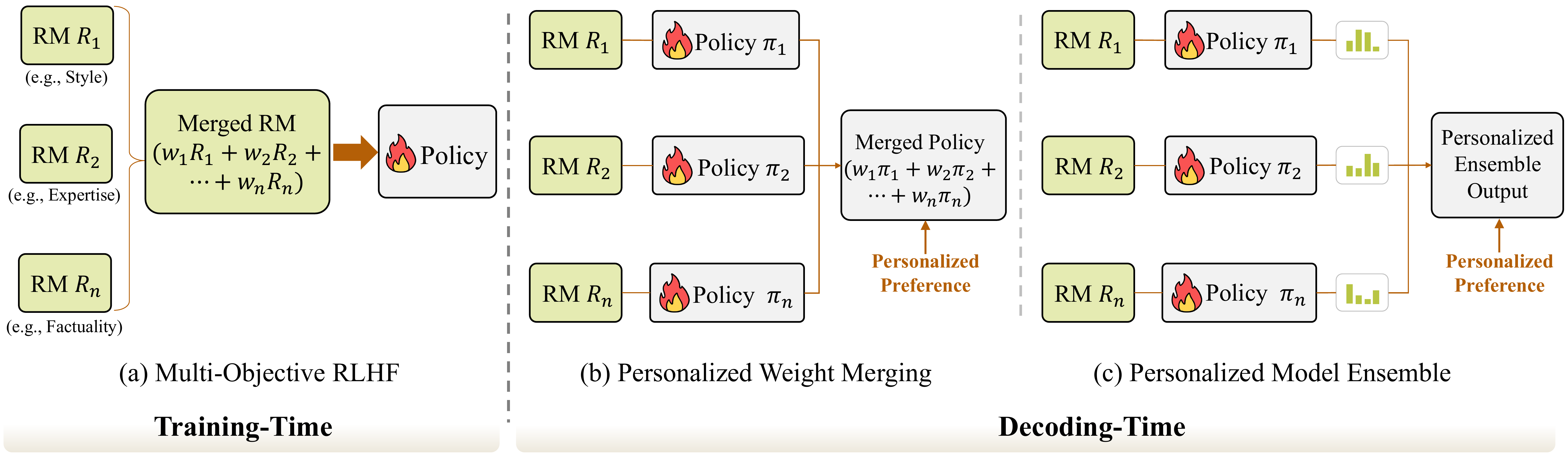}
    \caption{The illustration of personalized alignment approaches under the multi-objective reinforcement learning paradigm. \textbf{(a) Multi-Objective RLHF} trains separate reward models for different preference dimensions and uses their weighted combination to guide policy optimization. \textbf{(b) Personalized Weight Merging} combines multiple independently trained policy models with user-specific weights during inference. \textbf{(c) Personalized Model Ensemble} generates personalized outputs by ensembling predictions from multiple specialized policy models with dynamic user-specific weighting.}
    \label{fig:personalized_alignment}
\end{figure*}

\subsection{Personalized Alignment Optimization}
\label{subsec:optimization}

\begin{table}[!t]
    \centering
    \renewcommand\arraystretch{1.3}
    \small
    \caption{Comparison of Training-Time and Decoding-Time Personalization Alignment.}
    \label{tab:personalization_comparison}
    \vspace{4pt}
    \resizebox{\textwidth}{!}{
        \begin{tabular}{lll}
        \toprule[1.3pt]
        & \parbox[c]{6cm}{\centering \textbf{Training-Time Personalization}} & \parbox[c]{6cm}{\centering \textbf{Decoding-Time Personalization}} \\
        \midrule[1.3pt]
        
        \textbf{Training Stage} & 
        \parbox[c]{6cm}{\centering $\displaystyle \theta^* = \underset{\theta}{\mathrm{argmax}} \, \mathbb{E} \left[ \sum_i w_i R_i(x,y) \right]$} & 
        \parbox[c]{6cm}{\centering $\displaystyle \theta_i^* = \underset{\theta_i}{\mathrm{argmax}} \, \mathbb{E} [ R_i(x,y) ]$ \\ (for each $i$ in $1...n$)} \\
        
        \arrayrulecolor{gray} 
        \midrule[0.6pt]

        \textbf{Inference Stage} & 
        \parbox[c]{6cm}{\centering $y^* \sim \pi_{\theta^*}(x)$} & 
        \parbox[c]{6cm}{\centering 
            $\theta_p^* = \sum_i w_i \theta_i^*, y^* \sim \pi_{\theta_p^*}(x)$ (Merging) \\
            $y^* \sim \text{Decode}(\sum_i w_i P_{\theta_i^*})$ (Ensemble)
        } \\
        \arrayrulecolor{black} 
        \midrule[1.0pt]

        \rowcolor{mygreen2!35}
        \textbf{Pros.} & 
        \parbox[c]{6cm}{\centering {\color{darkgreen}\textbf{Strong Personalization}} \\ {\color{darkgreen}\textbf{Efficient Inference}}} & 
        \parbox[c]{6cm}{\centering {\color{darkgreen}\textbf{High Flexibility}} \\ {\color{darkgreen}\textbf{No Retraining}}} \\

        \arrayrulecolor{gray} 
        \midrule[0.6pt]
        
        \rowcolor{mygrey!20}
        \textbf{Cons.} &
        \parbox[c]{6cm}{\centering {\color{darkgray!90}\textbf{High Training Cost}} \\ {\color{darkgray!90}\textbf{Lack Flexibility}}} & 
        \parbox[c]{6cm}{\centering {\color{darkgray!90}\textbf{Storage Overhead}} \\ {\color{darkgray!90}\textbf{High Inference Cost}}} \\
        \arrayrulecolor{black} 

        \bottomrule[1.3pt]
        \end{tabular}
    }
\end{table}

Personalized preference alignment is usually modeled as a multi-objective reinforcement learning (MORL) problem, where personalized preference is determined as the user-specific combination of multi-preference dimensions. Based on this, a typical alignment paradigm is \textbf{training-time personalization}. This involves using a personalized reward derived from multiple reward models to guide the training of policy LLMs, which is illustrated on the left of \autoref{fig:personalized_alignment}.  \textbit{MORLHF}~\citep{wu2023fine} separately trains reward models for each dimension and retrains the policy language models using proximal policy optimization, guided by a linear combination of these multiple reward models. This approach allows for the reuse of the standard RLHF pipeline. \textbit{MODPO}~\citep{zhou2023beyond} introduces a novel RL-free algorithm extending Direct Preference Optimization (DPO) for managing multiple alignment objectives. 
It integrates linear scalarization into the reward modeling process, enabling the training of LMs using a margin-based cross-entropy loss as implicit collective reward functions.

The alternative paradigm is \textbf{decoding-time personalization}, which combines multiple trained policy LLMs during inference. \autoref{tab:personalization_comparison} provides a detailed comparison of this approach against training-time personalization. Its main strategies, \textbf{personalized weight merging} and \textbf{personalized model ensemble}, are illustrated in \autoref{fig:personalized_alignment}(b) and (c), respectively.
\textbit{Personalized Soups}~\citep{jang2023personalized} 
and \textbit{Reward Soups}~\citep{rame2024rewarded} address the challenge of RL from personalized human feedback by first training multiple policy models with distinct preferences independently and then merging their parameters post-hoc during inference. Both methods allow for dynamic weighting of the networks based on user preferences, enhancing model alignment and reducing reward misspecification.  Also, the personalized fusion of policy LLMs can be achieved not only through parameter merging but also through model ensembling.  \textbit{MOD}~\citep{shi2024decoding}  outputs the next token from a linear combination of all base models, allowing for precise control over different objectives by combining their predictions without the need for retraining. The method demonstrates significant effectiveness when compared to the parameter-merging baseline. \textbit{PAD}~\citep{chen2024pad} 
leverages a personalized reward modeling strategy to generate token-level rewards that guide the decoding process, enabling the dynamic adaptation of the base model’s predictions to individual preferences.

However, these approaches are limited by their reliance on predefined personalization dimensions and a small set of user preferences, which fail to capture the diversity of real-world personalization needs. Recent work has therefore sought to \textbf{extend reward learning to better accommodate user-specific preferences}. For example, \textbit{VPL}~\citep{poddar2024personalizing} employs a variational encoder to encode a small number of preference annotations from a given user into a latent variable that captures individual taste. A reward model is then conditioned on the latent variable and trained by maximizing the evidence lower bound, enabling accurate prediction of rewards aligned with the user’s unique preferences. \textbit{PREF}~\citep{shenfeld2025language} models each user’s personalized reward function as a linear combination of shared base reward functions, which are first learned from multi-user data via matrix factorization. Using an active learning strategy, it can efficiently infer a new user’s combination weights with only a handful of feedback examples (e.g., 10), enabling user-specific adaptation of LLMs' outputs without retraining. \textbit{PPT}~\citep{lau2024personalized} leverages in-context learning for scalable personalization by generating two candidate responses for each user prompt, soliciting user rankings, and dynamically incorporating this feedback into the model’s context to adapt to individual preferences over time.  \textbit{Drift}~\citep{kim2025drift} proposes a training-free method where user preference is modeled as a linear combination of interpretable attributes, and then computes a reward signal by comparing log-likelihoods under attribute-specific prompts to guide the generation of a frozen base model. In contrast to these offline methods based on static preferences, other approaches focus on real-time adaptation. \textbit{AMULET}~\citep{zhang2025amulet} approaches this from another unique test-time perspective, by formulating the decoding process of each token as an independent online learning problem. It obtains the optimization direction by contrasting the model's output with and without user-provided prompts and utilizes an efficient closed-form solution for real-time iterative optimization, thus adapting to personalized user needs instantly without retraining the model. \textbit{RLPA}~\citep{zhao2025teaching} simulates real users via a dynamically updated user portrait model, enabling the language model agent to engage in multi-turn interactive reinforcement learning within this simulated environment, thus continuously adapting its policy to evolving user preferences.  \textbit{PersonaAgent}~\citep{zhang2025personaagent} extends personalization from text style to decision behaviors such as tool usage, guided by a dynamic Persona that directs the agent’s decisions. Using test-time user-preference alignment, it compares the agent’s responses with the user’s real actions in real time, using textual loss as feedback to optimize the Persona, which ensures personalization run through every step of the tool decision.

\subsection{Discussions}
Current mainstream personalized alignment technologies mainly model personalization as multi-objective reinforcement learning problems, where personalized user preferences are taken into account during the training phase of policy LLMs via canonical RLHF, or the decoding phase of policy LLM via parameter merging or model ensembling.  Typically, these methods are limited to a small number (e.g., three) of predefined preference dimensions, represented through textual user preference prompts. However, in real-world scenarios, there could be a large number of personalized users, and their preference vectors may not be known, with only their interaction history accessed. Consequently, developing more realistic alignment benchmarks to effectively assess these techniques is a critical area for future research.

\newcommand{\multimodal}{{\color{deeppink}\ding{96}}}
\section{Metric and Benchmark}
\label{sec: benchmark}

As mentioned in \autoref{sec: problem statement}, LLM personalization can be categorized into multiple types based on different inputs, query types, and tasks. Currently, numerous benchmarks are available to help us validate personalization approaches. This section primarily organizes these benchmarks according to the aforementioned aspects. We first list the commonly used metrics for various personalization tasks (\autoref{subsec: metric}), then review the different benchmarks and the metrics they employ (\autoref{subsec: benchmark}).

\subsection{Metric} \label{subsec: metric}
We employ a comprehensive set of mainstream evaluation metrics to assess model performance across different tasks. The detailed specifications of these metrics are presented in \autoref{tab:metrics_specs}. We summarize Accuracy (Acc), F1 Score (F1), Matthews Correlation Coefficient (MCC)~\citep{chicco2020advantages}, Mean Absolute Error (MAE), and Root Mean Squared Error (RMSE) for the classification task; ROUGE-1 (R-1)~\citep{lin2004rouge}, BLEU~\citep{papineni2002bleu}, ROUGE-L (R-L)~\citep{lin2004rouge}, METEOR (MTR)~\citep{banerjee2005meteor}, SBERT~\citep{reimers2019sentence}, LLM-as-Evaluator (LLM-E)~\citep{gu2024survey} metrics to evaluate the quality of generated text for the generation task; and Hit Ratio (HR), Precision, Recall, and NDCG for the recommendation task.

\begin{table}[t]
\small
\centering
\renewcommand{\arraystretch}{2.2}  
\caption{Specifications of evaluation metrics used in our experiments. For each metric, we indicate the corresponding task type, and mathematical formulation. Notes indicate the applicable scenarios and characteristics of this metric. Three task types are included: Generation \textbf{G}, Classification \textbf{C}, and Recommendation \textbf{R}, where \Classification(M) and \Classification(O) indicate multi-class / binary-class and ordinal class, respectively.}
\label{tab:metrics_specs}
\rowcolors{2}{white}{mygrey!15}  
\begin{tabular}{cccc}
\toprule[1.3pt]
\textbf{Metric} & \textbf{Task Type} & \textbf{Formulation} & \textbf{Notes} \\
\midrule[1pt]
Accuracy (Acc) & \Classification (M)  & $\frac{TP + TN}{TP + TN + FP + FN}$ & Overall correctness \\
F1 & \Classification (M) & $2 \times \frac{precision \times recall}{precision + recall}$ & Precision and recall \\
\parbox[c]{4cm}{\centering MCC \\ \citep{chicco2020advantages}} & \Classification (M) &  $\frac{TP \cdot TN - FP \cdot FN}{\sqrt{(TP+FP)(TP+FN)(TN+FP)(TN+FN)}}$ &  Class imbalance \\
\parbox[c]{4cm}{\centering MAE \\ \citep{chai2014root}}  &  \Classification (O) & $\frac{1}{n}\sum_{i=1}^{n}|y_i - \hat{y}_i|$ & Outliers \\
\parbox[c]{4cm}{\centering RMSE \\ \citep{hodson2022root}}  & \Classification (O)  & $\sqrt{\frac{1}{n}\sum_{i=1}^{n}(y_i - \hat{y}_i)^2}$ & Large deviation penalty \\
\midrule
\parbox[c]{4cm}{\centering Perplexity \\ \citep{serban2016building}} & \Generation & $\exp\left( -\dfrac{1}{|R|} \sum_{i=1}^{|R|} \log p(w_i | \mathbf{w}_{<i}) \right)$ & Model's uncertainty \\
\parbox[c]{4cm}{\centering Rouge-1 (R-1) \\ \citep{lin2004rouge}} & \Generation & $\frac{|S \cap R|}{|R|}$ & N-gram overlap \\
\parbox[c]{4cm}{\centering BLEU \\ \citep{papineni2002bleu}} & \Generation & $\begin{cases} 
\frac{|S \cap R|}{|S|} \cdot \exp\left(1 - \frac{r}{c}\right) & \text{if } c < r \\
\frac{|S \cap R|}{|S|} & \text{if } c \geq r 
\end{cases}$ & Short length penalty \\
\parbox[c]{4cm}{\centering Rouge-L (R-L) \\ \citep{lin2004rouge}} & \Generation & $\frac{LCS(S,R)}{|R|}$ & Longest common subsequence \\
\parbox[c]{4cm}{\centering METEOR (MTR) \\ \citep{banerjee2005meteor}} & \Generation & $\frac{10m}{c + 9r} \cdot \left(1 - 0.5 \cdot \left(\frac{ch}{m}\right)^3\right)$ &  Linguistic features  \\
\parbox[c]{4cm}{\centering SBERT \\ \citep{reimers2019sentence,zhangbertscore}} & \Generation & $\cos(\mathbf{v}_S, \mathbf{v}_R) = \frac{\mathbf{v}_S \cdot \mathbf{v}_R}{\|\mathbf{v}_S\| \cdot \|\mathbf{v}_R\|}$ & Vector-based signal \\
\parbox[c]{4cm}{\centering LLM-as-Evaluator (LLM-E) \\ \citep{gu2024survey}} & \Generation & $\operatorname{LLM}(S, R)$ & Automated evaluation \\

\parbox[c]{4cm}{\centering EGISES \\ \citep{vansh2023accuracy}} & \Generation & $1 - \frac{1}{|\mathcal{C}| \cdot N} \sum_{c \in \mathcal{C}} \sum_{u_i \in \mathcal{U}} \text{Dev}(S_{c,u_i}, R_{c,u_i})$ & Personalization insensitivity \\

\parbox[c]{4cm}{\centering P-Accuracy \\ \citep{vansh2023accuracy}} & \Generation & $\text{Acc}(M) \cdot \left[1 - \alpha \cdot \frac{\sigma (\beta \cdot \text{EGISES}(M))}{\text{Acc}(M)}\right]$ & Personalization-aware accuracy \\

\parbox[c]{4cm}{\centering PerSEval \\ \citep{dasgupta2024perseval}} & \Generation & $\text{DEGRESS}(S, R) \times \text{EDP}(S, R)$ & Personalization evaluation \\

\midrule[1pt]

\parbox[c]{4cm}{\centering Hit Ratio (HR) } & \Recommendation & $\mathbf{1}\left(\sum_{i=1}^K rel_i > 0\right)$ & At least one hit \\ 

\parbox[c]{4cm}{\centering Precision \\ \citep{gunawardana2009survey}} & \Recommendation & $\frac{1}{K} \sum_{i=1}^K rel_i$ & Proportion of relevant items \\

\parbox[c]{4cm}{\centering Recall \\ \citep{gunawardana2009survey}} & \Recommendation & $\frac{\sum_{i=1}^K rel_i}{|\mathcal{G}|}$ & Coverage of user's relevant items \\

\parbox[c]{4cm}{\centering NDCG \\ \citep{wang2013theoretical}} & \Recommendation & $\frac{\sum_{i=1}^K \frac{rel_i}{\log_2(i+1)}}{\sum_{i=1}^{|R_K|} \frac{rel_i^*}{\log_2(i+1)}} $ & Position-aware relevance metric \\

\bottomrule[1.3pt] 
\end{tabular}
\end{table}

\subsubsection{Classification Task}

For the classification task, $TP$, $TN$, $FP$, and $FN$ represent true positives, true negatives, false positives, and false negatives, respectively; $y_i$ and $\hat{y}_i$ denote the ground truth and predicted values. 

\paragraph{Multi-class \& Binary Classification (M)}
In multi-class and binary classification tasks, where labels are categorical without inherent order, standard evaluation metrics such as Accuracy and F1 score are commonly employed to assess the model's ability to correctly predict class membership. MCC provides a more robust assessment under
class imbalance.

\paragraph{Ordinal Classification (O)} For ordinal multi-class classification, where labels possess a natural order or ranking, performance metrics like MAE and RMSE are preferred, as they account for the magnitude of prediction errors relative to the true order, providing a more nuanced evaluation of model quality.  RMSE reflects the square-root of the mean of squared errors,
providing a magnitude-sensitive measure that penalizes larger deviations more heavily. MAE
complements this by computing the average absolute error, offering robustness to outliers.

\subsubsection{Generation Task}

For the generation task, $S$ and $R$ represent the generated and reference sequences, and $LCS(S,R)$ indicates the length of the longest common subsequence between $S$ and $R$. $|S|$ and $|R|$ denote the number of unique words in the candidate and reference texts, respectively. In the Perplexity formula, $w_i$ is the $i$-th token in the reference text, $\mathbf{w}_{<i}$ represents all tokens that come before position $i$, and $p(w_i | \mathbf{w}_{<i})$ is the probability that the model assigns to token $w_i$ given the preceding tokens. In the BLEU formula, the brevity penalty is based on $c$ and $s$, which are the total word counts (including repeated words) in the candidate and reference sentences, respectively. In the METEOR formula, $m$ represents the number of matched words between the candidate text and reference text (including exact matches, stem matches, and synonym matches). $ch$ represents the number of "chunks" in the matching sequence, where a chunk is a contiguous sequence of matched words. A higher number of chunks indicates less fluent or more fragmented matching between texts. 

\paragraph{Human-based Evaluation (Human-E)} Human-E is a process where human judges assess LLM outputs by rating them on accuracy, helpfulness, coherence, appropriateness, etc. In this approach, human evaluators review and rate responses based on criteria such as accuracy, helpfulness, coherence, and appropriateness. Human-E is considered the gold standard because humans can detect nuances in language, recognize factual errors, understand contextual appropriateness, and make holistic judgments that automated metrics often miss.

\paragraph{Conventional Evaluation} BLEU, ROUGE-1, ROUGE-L, and METEOR to measure lexical overlap between the generated $y$ and ground-truth $\hat{y}$ responses. 
BLEU evaluates n-gram precision focusing on system output, whereas ROUGE-1 measures unigram recall relative to reference texts. ROUGE-L extends this by capturing the longest common subsequence structure. METEOR incorporates linguistic features, including stemming and synonym matching, for enhanced evaluation. For semantic-level assessment beyond surface-level patterns, {\color{black} SBERT-based cosine similarity~\citep{li2025personalized} provides a vector-based measure of meaning preservation between generated and reference responses.}

\paragraph{LLM-based Evaluation (LLM-E)}   Human-based evaluation provides the most reliable assessment of LLM response quality, but obtaining statistically significant results is costly and time-intensive. Therefore, the “LLM-as-a-judge”  framework~\citep{gu2024survey} is proposed that uses LLM to automatically evaluate the quality and relevance of generated text. By prompting LLMs to score or compare outputs, it offers scalable, context-aware, and semantically rich assessments with minimal human input. This approach surpasses traditional metrics in flexibility but faces challenges like model bias and consistency. It represents a promising method for automated, nuanced evaluation. ExPerT~\citep{salemi2025expert} introduces an explainable reference-based evaluation framework specifically designed for style-based personalized text generation. It leverages an LLM to extract atomic aspects and their evidence from the generated and reference texts, match the aspects, and evaluate their alignment based on content and writing style -- two key attributes in personalized text generation.

\paragraph{Personalization Degree Evaluation (Per-E)} 
The aforementioned metrics are unable to assess the personalization capability of summarization models. To address this limitation, EGISES~\citep{vansh2023accuracy} is proposed as the first automatic measure for evaluating personalization in text summarization. The core Deviation function measures summary-level personalization by computing the ratio $\frac{\min(X_{u,u'}, Y_{u,u'})}{\max(X_{u,u'}, Y_{u,u'})}$ averaged across all user pairs, where $X_{u,u'}$ and $Y_{u,u'}$ represent the Jensen-Shannon divergence-weighted differences between users' expected summaries and model-generated summaries, respectively. 
The Deviation function measures personalization quality: $\text{Dev}(S_{c,u_i}, R_{c,u_i}) \rightarrow 1$ for good personalization, $\text{Dev}(S_{c,u_i}, R_{c,u_i}) \rightarrow 0$ for poor personalization, where the model fails to match user expectation differences. Building upon EGISES, P-Accuracy~\citep{vansh2023accuracy} provides a personalization-aware accuracy measure that penalizes traditional accuracy scores based on the model's personalization capability,  $\text{Acc}(M) \cdot [1 - \alpha \cdot \frac{\sigma (\beta \cdot \text{EGISES}(M))}{\text{Acc}(M)}]$, where $\sigma$ is the sigmoid function, $\alpha$ and $\beta$ control the penalty intensity and personalization emphasis, respectively. However, EGISES suffers from the \textit{personalization-accuracy paradox}, where models can achieve high responsiveness but low accuracy, leading to poor user experience. PerSEval~\citep{dasgupta2024perseval} addresses this limitation by introducing the EDP that penalizes EGISES scores when accuracy drops, computed as $\text{PerSEval} = \text{DEGRESS} \times \text{EDP}$, where DEGRESS measures responsiveness, and EDP incorporates accuracy-based penalties to ensure that personalization evaluation considers both user preference alignment and content quality.

\subsubsection{Recommendation Task}

For the recommendation task, $rel_i$ denotes the relevance score of the item at rank position $i$, which may be binary (0 or 1) or graded (multi-level) depending on the evaluation context.
In the NDCG formula, $|R_K|$ is the number of relevant items (limited to $K$ items),
$rel_i^*$ represents the relevance values sorted in descending order (the ideal ordering).

\paragraph{Item Recommendation}
Traditional recommendation tasks typically use Hit Ratio (HR), Recall, and Discounted Cumulative Gain (NDCG) ~\citep{ning2024user, ramos2024preference} as standard evaluation metrics to measure the effectiveness of top-K recommendation and preference ranking.

\paragraph{Conversational Recommendation} Conversational recommendation systems commonly use Recall and NDCG as evaluation metrics to measure coverage and ranking quality, employing the "LLMs-as-a-judger" framework
~\citep{zhao2025exploring, huang2024concept, sayana2024beyond}. Additionally, an LLM-based user simulator—creating unique personas via zero-shot ChatGPT prompting and defining preferences using dataset attributes—is also used to assess whether outputs align with user preferences~\citep{huang2024concept}. \textbf{Note that} in this scenario, we still categorize it as a generation task (\textbf{G}), as the core objective remains text generation in context, while the input data contains interaction information.

\definecolor{deeppink}{HTML}{E75480}   

\begin{table*}[htbp]
    \centering
    \small
    \renewcommand{\arraystretch}{2.35} 
    \caption{
Summary of personalized large language model benchmarks. Five data categories are shown: Historical \textbf{Content}, \textbf{Dialogues}, Interactions, \textbf{User Profile}, and Pre-defined Human \textbf{Preference} (\autoref{sec: problem statement}). Three task types are included: Generation \textbf{G}, Classification \textbf{C}, and Recommendation \textbf{R}. Techniques (abbreviated as): Prompting \textbf{PT} (\autoref{sec: personalized prompting}), Adaptation \textbf{AD}~(\autoref{sec: personalized adatation}), and Alignment \textbf{AL}~(\autoref{sec: personalized alignment}). Three query types are included: Abstraction, Extraction and Generalization. \textbf{Note:} In the row for 'All (excl. LLM-E)', 'All' refers to all metrics corresponding to the tasks listed in ~\autoref{tab:metrics_specs}; '\textit{excl.}' abbreviates 'excluding'. * indicates methods built following the benchmark rather than methods from the benchmark. \multimodal indicates the benchmark comprises multi-modal personalized data or output.}

\rowcolors{2}{white}{mygrey!15}  
    \label{tab:personalized_benchmarks}
\resizebox{\textwidth}{!}{
    \begin{tabular}{cccccc}
        \toprule[1.3pt]
        \textbf{Benchmark} & \textbf{Personalized Data} & \textbf{Query} & \textbf{Task} & \textbf{Technique} & \textbf{Metric} \\
        \midrule[1.3pt]
        \rowcolor{mygrey!15}

        \parbox[c]{2.5cm}{\centering MemoryBank \\ \citep{xue2025mmrc}} & \parbox[c]{2cm}{\centering User Profile \\ Dialogues} & Extraction & \Generation & \centering \textbf{PT} & LLM-E  \\

        \parbox[c]{2.5cm}{\centering PerLTQA \\ \citep{du2024perltqa}} & \parbox[c]{2cm}{\centering User Profile \\ Dialogues} & Extraction & \Generation & \centering \textbf{PT} (RAG) & LLM-E  \\

        \parbox[c]{2.5cm}{\centering LoCoMo \multimodal \\ \citep{maharana2024evaluating}} & \parbox[c]{2cm}{\centering User Profile \\ Dialogues} & \parbox[c]{2cm}{\centering Abstraction \\ Extraction \\ Generalization} & \Generation & \centering \textbf{PT} (RAG) & F1, Recall  \\
        
        \parbox[c]{2.5cm}{\centering LongMemEval \\ \citep{wulongmemeval}} & Dialogues & Extraction & \Generation & \textbf{PT} (RAG) & LLM-E  \\
        
        \parbox[c]{2.5cm}{\centering MMRC \multimodal \\ \citep{xue2025mmrc}} & Dialogues & Extraction & \Generation & \textbf{PT} & LLM-E, Human-E  \\

        \parbox[c]{2.5cm}{\centering IMPLEXCONV \\ \citep{li2025toward} } & Dialogues & \makecell[c]{Extraction \\ Generalization} & \Generation & \textbf{PT} (RAG) & LLM-E  \\
        
        \parbox[c]{2.5cm}{\centering PrefEval \\ \citep{zhao2025llms}} & Dialogues & \parbox[c]{2cm}{Generalization} &  \Generation, \Classification & \parbox[c]{2cm}{\centering \textbf{PT} (RAG) \\ SFT} & \parbox[c]{4cm}{\centering Acc, LLM-E }  \\

        \parbox[c]{3.5cm}{\centering HiCUPID \\ \citep{mok2025exploring}} & \parbox[c]{2cm}{\centering Preference
        \\ Dialogues } & Extraction & \textbf{G} & \parbox[c]{2cm}{\centering \textbf{PT} (RAG), SFT \\ \textbf{AL} (DPO)} & BLEU, R-L, LLM-E  \\

        \parbox[c]{2.5cm}{\centering {\color{black} MemBench} \\ \citep{tan2025membench}} & Dialogues & \parbox[c]{2cm}{\centering Abstraction \\ Extraction \\ Generalization}  &  \Generation & \parbox[c]{2cm}{\centering \textbf{PT} (RAG)} & \parbox[c]{4cm}{\centering Acc }  \\

        \midrule[1pt]

        \parbox[c]{2.5cm}{\centering PER-CHAT \\ ~\citep{wu2021personalized}} & \parbox[c]{2cm}{\centering User Profile \\ Content} & Extraction & \Generation & \textbf{PT} (Soft) &  \parbox[c]{4cm}{\centering Human-E, Perplexity,\\ BLEU}  \\
        
        \parbox[c]{2.7cm}{\centering LaMP \\ \citep{salemi2023lamp}} & Content & \parbox[c]{2cm}{\centering Abstraction \\ Generalization}  & \parbox[c]{2cm}{\centering \Generation, \Classification} & \parbox[c]{1.9cm}{\centering \textbf{PT}, \textbf{AD}*, \textbf{AL}*} & \parbox[c]{2.5cm}{\centering Acc, F1, \\ MAE, RMSE, \\ R-1, R-L}  \\

        \parbox[c]{2.7cm}{\centering LongLaMP \\ \citep{kumar2024longlamp}} & Content & \parbox[c]{2cm}{\centering Generalization}  & \Generation & \textbf{PT} (RAG) & R-1, R-L, MTR \\

        \parbox[c]{3.5cm}{\centering PEFT-U \\ \citep{clarke2024peft}} & Content & \parbox[c]{2cm}{\centering Abstraction} & \Classification & \textbf{FT} & Acc  \\
        
        \parbox[c]{2.7cm}{\centering pGraphRAG \\ \citep{au2025personalized}} & Content & \parbox[c]{2cm}{\centering Abstraction \\ Generalization}  & \parbox[c]{1cm}{\centering \Generation, \Classification} & \textbf{PT} (RAG) & \parbox[c]{2cm}{\centering MAE, RMSE, \\ R-1, R-L, MTR}  \\

        \parbox[c]{3.5cm}{\centering LaMP-QA \\ \citep{salemi2025lamp}} & \parbox[c]{2cm}{\centering Content} & \parbox[c]{2cm}{\centering Generalization}  & \Generation & \parbox[c]{2cm}{\centering \textbf{PT} \\ (PAG + RAG)} & \parbox[c]{2cm}{\centering 
        LLM-E}  \\
        
         \parbox[c]{3.5cm}{\centering DPL \\ \citep{qiu2025measuring}} & Content & \parbox[c]{2cm}{\centering Generalization} & \Generation & \textbf{PT} (PAG) & \parbox[c]{3cm}{\centering R-1, R-L, BLEU, \\ MTR, LLM-E}   \\
         
        \parbox[c]{2.5cm}{\centering PERSONABench \\ \citep{li2025personalized}} & Content & \parbox[c]{2cm}{\centering Abstraction \\ Generalization} &  \Generation, \Classification & \textbf{PT} (ICL) & \parbox[c]{4cm}{\centering All (\textit{excl.} LLM-E, Per-E)} \\

        \midrule[1pt]

        \parbox[c]{3.5cm}{\centering PRISM \\ \citep{kirk2024prism}} & Preference & Generalization & \Generation  & \textbf{AL}* &  -- \\
        
        \parbox[c]{3.5cm}{\centering
        PersonalLLM \\ \citep{zollo2024personalllm}} & Preference & Generalization & \Generation & \parbox[c]{1.8cm}{\centering \textbf{PT} (ICL)} & LLM-E  \\

        \parbox[c]{3.5cm}{\centering ALOE \\ \citep{wu2024aligning}} & \parbox[c]{2cm}{\centering User Profile \\ Preference} & Generalization & \textbf{G} & \parbox[c]{1.8cm}{\centering \textbf{AL} (DPO)\\SFT} & LLM-E  \\

        \midrule[1pt]

        \parbox[c]{3.5cm}{\centering REGEN \\ \citep{sayana2024beyond}}& Interactions & Generalization & \textbf{G} & \textbf{PT} (Soft) & \parbox[c]{3cm}{\centering R-L, BLEU, \\ SBERT}  \\

        \parbox[c]{3.5cm}{\centering 
        PersonalWAB \\ \citep{cai2025large}} & \parbox[c]{2cm}{\centering User Profile \\ Interactions}  & Generalization & \textbf{R} & \parbox[c]{2cm}{\centering \textbf{PT + AL} \\ (RAG + \\ SFT / DPO)} & Acc  \\

        \parbox[c]{3.5cm}{\centering 
        RecBench+ \\ \citep{huang2025towards}} & \parbox[c]{2cm}{\centering User Profile \\ Interactions}  & \parbox[c]{2cm}{\centering Extraction \\ Generalization} & \textbf{R} & -- & \parbox[c]{3cm}{\centering Precision, Recall
        } \\

        \bottomrule[1.3pt]
    \end{tabular}
}
\end{table*}

\subsection{Benchmark} \label{subsec: benchmark}

Benchmarks for LLM personalization can be categorized based on the types of input data they utilize, the nature of the queries they address, and the specific tasks they are designed to evaluate. In terms of the personalized data format, benchmarks can be classified into four main types: Historical Content, Dialogues, Interactions, and Pre-defined Human Preference. Sometimes the user profiles are also included as auxiliary information. The mainstream benchmarks for each data type are summarized in \autoref{tab:personalized_benchmarks}~\footnote{PERSONABench in ~\ref{tab:personalized_benchmarks} is the abbreviation for PERSONACONVBENCH~\citep{li2025personalized}.}\footnote{Among fact-based query benchmarks (such as PerLTQA~\citep{du2024perltqa}, LongMemEval~\citep{wulongmemeval}, and IMPLEXCONV~\cite{li2025toward}), some include memory-related testing tasks, like retrieval. Since these are not the focus of this paper, which investigates techniques for injecting and utilizing personalized user data to guide LLMs toward better personalized performance, these tasks are not listed in the summary.}.

\paragraph{Dialogue-based Data} For dialogue-based data, the benchmarks typically focus on the ability of LLMs to extract and utilize personalized information (e.g., user characteristics, schedules, preferences) from multi-turn conversations. The challenging aspect here is to \textbf{accurately capture user intent and context from the long-term dialogue history}, which may involve complex interactions, implicit preferences, and enormous unrelated content~\citep{li2025toward, zhao2025llms, mok2025exploring}. Therefore, the mainstream techniques for this type of data are prompting-based methods, which can effectively leverage the most decisive information from the dialogue history without requiring extensive model retraining. The query types in these benchmarks often include explicit extraction of personalized factual details. Naturally, the metrics used are usually accuracy-based, such as F1 score and accuracy, with LLM-E as a preprocessing step to calculate the exact matching scores between the generated and ground-truth responses~\citep{xue2025mmrc, wulongmemeval, tan2025membench}.

\paragraph{Historical Content-based Data} For historical content-based data, the benchmarks often evaluate how well LLMs can incorporate and reason over user-specific documents, such as past tweets, articles, or other textual materials. The key challenge is to enable the model to \textbf{understand and integrate this external knowledge into its responses}, which may require sophisticated retrieval and comprehension capabilities~\citep{salemi2023lamp, kumar2024longlamp}. Benefiting from the purer text content provided in these benchmarks, techniques like SFT are more commonly employed to intrinsically adapt the model to the user's content, while the prompting-based RAG and PAG methods are auxiliarily used to further retrieve and conclude the most relevant terms for fine-tuning~\citep{au2025personalized, salemi2025lamp}. More query types of abstraction and generalization are included in these benchmarks, which require the model to not only extract factual information but also adapt its style and tone to match user preferences~\citep{salemi2023lamp}. Therefore, the metrics used are more diverse, including both accuracy-based metrics and generation quality metrics like ROUGE, BLEU, and METEOR, with LLM-E and Human-E as supplementary evaluations~\citep{li2025personalized}.

\paragraph{Preference-based Data} For preference-based data, the benchmarks typically assess how well LLMs can align their outputs with user-defined preferences, which may be explicitly stated or implicitly inferred from user behavior. The main challenge is to \textbf{accurately model and incorporate these preferences into the generation process}, which may involve complex reasoning and decision-making~\citep{kirk2024prism, wu2024aligning}. Techniques like RLHF and DPO are often employed to fine-tune the model based on user feedback, while prompting-based methods like ICL are also used to guide the model's behavior during inference. Targeting the subjective nature of preferences, the query types in these benchmarks often involve open-ended generation tasks, where the model must produce responses that are not only factually accurate~\citep{mok2025exploring, zollo2024personalllm} but also align with user tastes and styles~\citep{kirk2024prism}. Consequently, the metrics used are primarily generation quality metrics, with LLM-E and Human-E as critical components to evaluate the alignment with user preferences.

\paragraph{Interaction-based Data} For interaction-based data, the benchmarks often focus on how well LLMs can leverage user interaction histories, such as clicks, likes, and other engagement signals, to personalize recommendations and responses. The key challenge is to \textbf{effectively model user behavior and preferences from these interactions}, which may be sparse and noisy~\citep{sayana2024beyond, huang2025towards}. Techniques like fine-tuning are commonly used to adapt the model to user interaction patterns, while prompting-based methods are also employed to retrieve relevant information based on past interactions~\citep{cai2025large}. The query types in these benchmarks often involve generalized recommendation tasks, where the model must generate personalized suggestions based on user behavior. Therefore, the metrics used are typically recommendation quality metrics, such as Hit Ratio and NDCG. Sometimes, the generation quality metrics like BLEU and ROUGE are also used when the recommendation is presented in a conversational format.

\subsection{Discussions}

Although existing benchmarks cover a wide range of data types, query types, and tasks, there are still some limitations and gaps that need to be addressed in future research. 
To generalize to more realistic scenarios, future benchmarks should consider the following aspects:
 
\begin{myinfobox}

{{\hspace{0.05cm}\raisebox{-0.35\height+0.25ex}{\includegraphics[width=0.6cm]{Figures/icons/issue.png}} \color{darkgreen}\textbf{LLM Personalization Benchmarks: Challenges and Directions}}}

  \paragraph{Lifelong Update} In real-world applications, user preferences and behaviors may change over time, requiring LLMs to continuously adapt and update their personalization strategies. Future benchmarks should include scenarios that test the model's ability to continuously \emph{update} knowledge, \emph{resolve} conflicts, and \emph{forget} outdated information.

  \paragraph{Cross-Domain Adaptation} Most existing benchmarks focus on a single domain or task, which may limit the generalizability of the personalization techniques. Future benchmarks should include scenarios that test the model's ability to \emph{adapt to different domains and tasks}, such as switching from news summarization to product recommendation.
  
  \paragraph{Cross-Modal Personalization} Most existing benchmarks focus on text-based data, which may not fully capture the richness and diversity of user preferences~\footnote{ Although some benchmarks like LoCoMo\multimodal ~\citep{maharana2024evaluating} and MMRC\multimodal ~\citep{xue2025mmrc} include multi-modal data, they are still limited in scale and scope.}. Future benchmarks should include scenarios that test the model's ability to \emph{integrate and utilize multi-modal data}, such as images, videos, and audio.

  \paragraph{Information Scarcity and Imbalance} Due to privacy and security concerns, collecting large-scale, high-quality personalized data for training and evaluation is often challenging. As a result, the gathered user data may be scarce, noisy, or imbalanced. Future benchmarks should incorporate more demanding scenarios to assess a model's ability to \emph{detect} inconsistencies, \emph{refuse} to answer questions beyond its knowledge, and \emph{generalize} effectively from limited or biased data.
  
\end{myinfobox}

\section{Vision and Future Directions}
\label{sec: future directions}

\subsection{Vision and Challenges}

\begin{figure*}[!t]
    \centering
    \includegraphics[width=0.99999\linewidth]{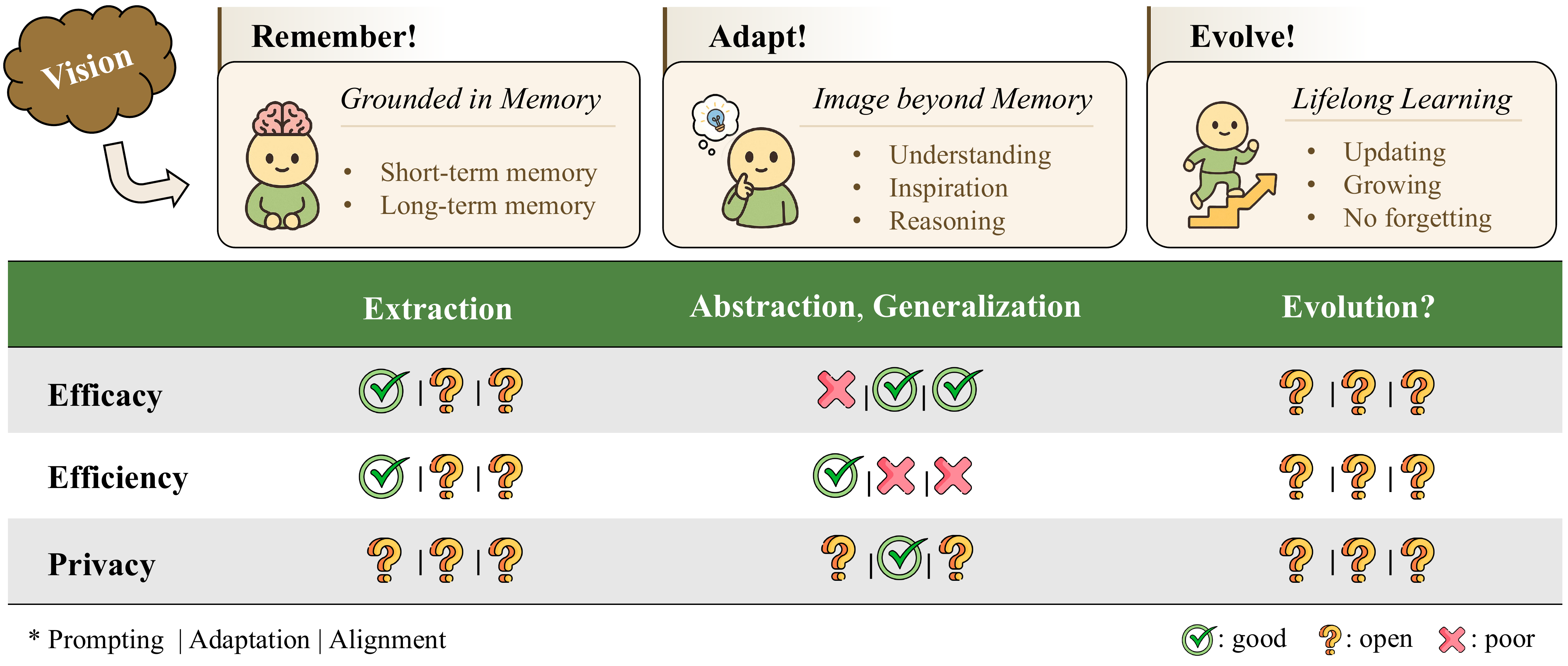}
    \caption{\textbf{Vision for PLLMs.}The figure contrasts three capability axes—\emph{Efficacy}, \emph{Efficiency}, and \emph{Trustworthiness}—across different dimensions: \emph{Extraction}, \emph{Abstraction / Generalization}, and \emph{Evolution}. Within each cell, icons summarize (left to right) \emph{Prompting ~(\autoref{sec: personalized prompting})} \textbar{} \emph{Adaptation ~(\autoref{sec: personalized adatation})} \textbar{} \emph{Alignment ~(\autoref{sec: personalized alignment})}. The meaning of Legend is as follows: \includegraphics[height=1.2em]{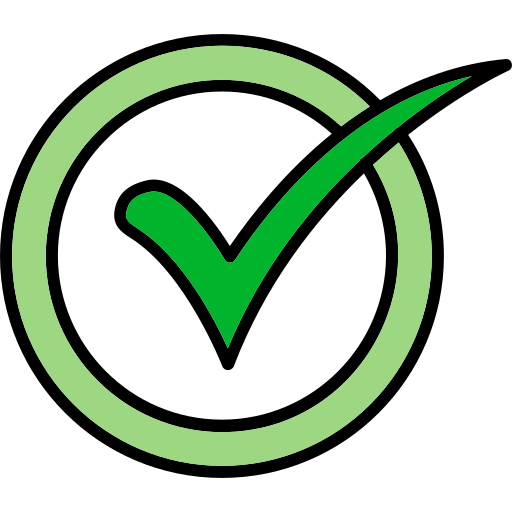}(good/strong, capability already well-demonstrated); 
\includegraphics[height=0.8em]{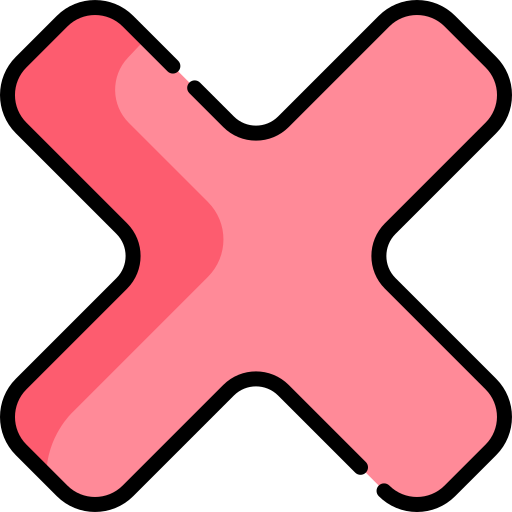} (poor/weak, capability not well-supported); 
\includegraphics[height=1em]{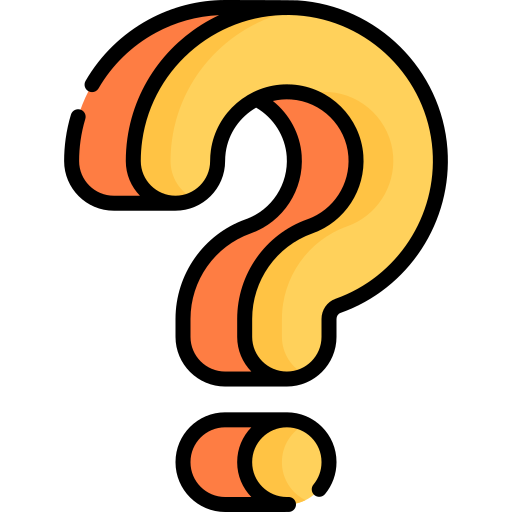} (open/under-explored, requiring further research). }
    \label{fig:summarization}
\end{figure*}

The vision of PLLMs (\autoref{fig:summarization}) is to endow models with the abilities to remember, adapt, and evolve. \textbf{Remembering} means grounding responses in user-specific knowledge from both short- and long-term memory. \textbf{Adapting} highlights the capacity to abstract and generalize from personalized data, enabling reasoning and dynamic adjustment to user preferences. \textbf{Evolving} emphasizes continual learning, where models update personalization without forgetting, where models update personalization without forgetting, staying aligned with users’ shifting goals and their growth over time.

We further analyze the vision of \emph{PLLMs} from the perspective of three key techniques: \emph{prompting}, \emph{adaptation}, and \emph{alignment}. 
As shown in \autoref{fig:summarization}, each technique contributes differently across the stages of \emph{extraction}, \emph{abstraction/generalization}, and \emph{evolution}, and each faces distinct trade-offs in terms of efficacy, efficiency, and privacy. 

\textbf{Efficacy} 
At the extraction task, prompting is effective and efficient, enabling PLLMs to recall and leverage factual or personalized information with minimal overhead~\citep{tan2025membench}. 
However, as tasks move toward abstraction and generalization, the limitations of prompting become evident, and adaptation/alignment methods play a more central role~\citep{tan2024personalized, bu2025personalized}.  
These techniques allow models to internalize user-specific preferences, improving personalization depth, but often come at higher computational and data requirements. 
Finally, at the evolution stage, where continual updating and lifelong learning are essential, performance remains largely under-explored, as models struggle to grow and adapt without catastrophic forgetting.  

\paragraph{Efficiency} 
Efficiency is a constant constraint across all goals. 
Extraction via prompting is highly efficient, but as adaptation and alignment become necessary for abstraction, efficiency drops sharply due to the cost of fine-tuning or maintaining additional modules~\citep{clarke2024peft}. 
At the evolution stage, efficiency challenges are amplified, since continual updating requires frequent synchronization between local devices and cloud servers, incurring latency, storage, and energy costs. 
This highlights a tension where methods that improve personalization depth often conflict with the need for scalable, lightweight deployment.  

\paragraph{Privacy} 
A key challenge is maintaining user privacy while still enabling personalization. 
Per-user PEFT strategies provide strong privacy guarantees but at the cost of reduced performance~\citep{peng2024pocketllm}. 
In contrast, collaborative or federated strategies can enhance performance by leveraging cross-user signals, but inevitably raise the risk of privacy leakage during communication or model aggregation~\citep{zhuang2024hydra}. 
Parameter-sharing approaches offer a compromise by avoiding direct exposure of raw user data, yet they shift the bottleneck to communication and synchronization efficiency in edge–cloud coordination~\citep{qi2024fdlora}.

\textbf{Overall Trade-offs} 
Taken together, the figure illustrates a triangular trade-off: methods that strongly guarantee privacy tend to limit performance; those that maximize performance often compromise privacy; and strategies that balance the two introduce efficiency bottlenecks. 
While prompting-based extraction already achieves good efficacy and efficiency, abstraction and especially lifelong evolution remain under-explored. 
Addressing this “trilemma” requires new techniques that can simultaneously enhance performance, protect privacy, and scale efficiently—a direction that we identify as a key frontier for future research.

\subsection{Future Directions}

Despite advances in Personalized Large Language Models (PLLMs), significant challenges persist, particularly in \textbf{technical improvements}. Current methods effectively handle basic user preferences but struggle with complex, multi-source data, especially in multimodal contexts like images and audio. Efficiently updating models on resource-constrained edge devices is also crucial. Fine-tuning enhances personalization but can be resource-intensive and difficult to scale. Developing small, personalized models through techniques like quantization could address these issues.

\begin{enumerate}
    \item \textbf{Complex User Data} While current approaches effectively handle basic user preferences, processing complex, multi-source user data remains a significant challenge. For example, methods that use user relationships in graph-like structures are still limited to retrieval augmentation ~\citep{du2024perltqa}. How to effectively leverage this complex user information to fine-tune LLM parameters remains a significant challenge. 
    Most methods focus on text data, while personalized foundation models for multimodal data (e.g., images, videos, audio) remain underexplored, despite their significance for real-world deployment and applications~\citep{wu2024personalized, pi2024personalized, shen2024pmg, xu2025personalized}.

    \item \textbf{Edge Computing} A key challenge in edge computing is efficiently updating models on resource-constrained devices (e.g., phones), where storage and computational resources are limited. For example, fine-tuning offers deeper personalization but is resource-intensive and hard to scale, especially in real-time applications. Balancing resources with personalization needs is important. 
    A potential solution is to build personalized small models~\citep{lu2024small} for edge devices, using techniques like quantization and distillation.

    \item \textbf{Edge-Cloud Collaboration} The deployment of PLLMs in real-world scenarios encounters significant challenges in edge-cloud computing environments. Current collaborative approaches often lack efficient synchronization between cloud and edge devices, highlighting the need to balance local computation and cloud processing ~\citep{tian2024edge}.

    \item \textbf{Efficient Adaptation to Model Updates} Updating fine-tuned PEFT parameters for each user when base LLM parameters change poses a challenge due to high user data volume and limited resources. Retraining costs can be prohibitive. Future research should focus on efficient methods for updating user-specific parameters without complete retraining, such as incremental learning and transfer learning. Moreover, the emerging paradigm of small language models (SLMs)~\citep{liu2024mobilellm} offers a promising direction, where lightweight models can be co-trained or specialized to handle personalization updates more efficiently while reducing memory and energy costs.

    \item \textbf{Lifelong Updating} Given the large variety of user behaviors, a key challenge is preventing catastrophic forgetting while ensuring the efficient update of long-term and short-term of memory. Future research could explore continual learning~\citep{wu2024continual} and knowledge editing~\citep{wang2024knowledge, zhang2024comprehensive} to facilitate dynamic updates of user-specific information.
\end{enumerate}

\textbf{Privacy} Privacy remains a critical concern, particularly regarding user privacy when generating personalized responses. As LLMs are not typically deployed locally, risks of privacy leakage arise. Future research should focus on privacy-preserving methods, such as federated learning and differential privacy~\citep{yao2024federated, liu2024client}, to protect user data effectively while leveraging the model's capabilities.

\textbf{Trustworthiness} Beyond performance and efficiency, ensuring trustworthiness is critical for the adoption of personalized LLMs. 
Future research should address issues of \emph{interpretability}, enabling users to understand why a personalized response is generated and to what extent it relies on their data~\citep{he2024cos}, like neuron selection~\citep{zhao2025sparse}.  
Another direction is improving \emph{fairness and bias mitigation}, since personalization risks amplifying stereotypes or reinforcing undesirable patterns if user-specific data is skewed. 
Ultimately, developing standardized evaluation protocols and benchmarks for trustworthiness in the context of personalization will be key to building systems that users not only find useful but can also confidently rely on. 

\textbf{Applications} Exploring domain-specific challenges and opportunities for personalization is essential to advance the practical adoption of PLLMs. For example, in \textit{healthcare}, it promises tailored treatments but must strictly safeguard privacy~\citep{zhang2024llm}. In \textit{education}, adaptive tutoring is enabled, yet fairness and long-term modeling remain concerns~\citep{zhao2025learnlens}. 
In \textit{creative industries}, they support style-aware generation while risking overfitting and reduced diversity. In \textit{enterprise applications}~\citep{li2025memos}, personalization enhances productivity and customer support, but requires scalable, consistent, and governable solutions.

\newcommand{\lamp}{{\color{orange}\ding{75}}}
\newcommand{\longlamp}{{\color{myblue}\ding{75}}}
\newcommand{\perltqa}{{\color{magenta}\ding{75}}}
\newcounter{quadcount}
\newcommand{\removeblank}[1]{%
  \setcounter{quadcount}{0}%
  \loop
    \ifnum\value{quadcount}<#1%
      \quad%
      \stepcounter{quadcount}%
    \repeat
}

\begin{table*}[htbp]
\renewcommand\arraystretch{1.2}
\caption{A systematic categorization of personalization strategies for PLLMs. Methods marked with {\lamp} use the LaMP benchmark~\citep{salemi2023lamp}, {\longlamp} use the LongLaMP benchmark~\citep{kumar2024longlamp}, and {\perltqa} use the PerLTQA benchmark~\citep{du2024perltqa}. (o) means optional. The overview presents four data categories (Historical Content, Dialogues, Interactions, User Profile, and Pre-defined Human Preference) and three task types (Generation \textbf{G}, Classification \textbf{C}, and Recommendation \textbf{R}), along with fine-tuning requirements for generator LLMs.}
\rowcolors{2}{white}{mygreen!30}  
\resizebox{\textwidth}{!}{
\begin{tabular}{lp{2.5cm}p{2cm}p{6cm}cccl}
\toprule[1.3pt]
    \textbf{Method} & \textbf{Personalized Data} & \textbf{Query} & \textbf{LLM (Generator)} & \textbf{Retriever} & \textbf{Prompting} & \textbf{Fine-tuning} & \textbf{Task} \\ \midrule[1pt]

    \rowcolor{gray!10} 
    \multicolumn{8}{l}{\color{black} \textsection\ref{sec: personalized prompting}\quad \textbf{Personalized Prompting}} \\ 
    \midrule

    Cue-CoT~\citep{wang2023cue} &   Dialogues & \makecell[l]{Genl.} & \makecell[l]{ChatGLM-6B, BELLE-LLaMA-7B-2M,\\ ChatGPT, Alpaca-7B, Vicuna-7B-v1.1} & $\times$ & Token & $\times$ &   \textbf{G}\\ %

    PAG~\citep{richardson2023integrating}~\lamp &   Content & \makecell[l]{Abs., Genl.} & GPT-3.5, Flan-T5, Vicuna-13B & $\times$ & Token & $\times$ &   \textbf{G},   \textbf{C} \\ %

    Matryoshka~\citep{li2024matryoshka}~\lamp &   Content & \makecell[l]{Abs., Genl.} & \makecell[l]{GPT-4o-mini, GPT-3.5-Turbo} & $\times$ & Token & $\times$ &   \textbf{G},   \textbf{C}\\ 

    RewriterSlRl~\citep{li2024learning} &   Content & \makecell[l]{Genl.} & PaLM-2 & $\times$ & Token & $\times$ &   \textbf{G} \\ 

    R2P~\citep{luo2025reasoning}~\lamp  & Content & \makecell[l]{Abs., Genl.} & LLaMA-3.1-8B-Instruct, DeepSeek-R1-Distill-LLaMA-8B & $\times$ & Token & $\times$ & \textbf{G}, \Classification \\ 

    DPL~\citep{qiu2025measuring} &  Content  & \makecell[l]{Genl.} & GTE-Qwen2-1.5B-Instruct & $\times$ & Token & $\times$ &   \textbf{G} \\ 

    CoS~\citep{he2024cos} & User Profile & \makecell[l]{Abs., Genl.} & \makecell[l]{ GPT-3.5, LLaMA-2-7B-Chat, T0pp, \\ Mistral-7B-Instruct} & $\times$ & Token & $\times$ &   \textbf{G},   \textbf{C} \\ 

    StyleVector~\citep{zhang2025personalized}~\lamp \longlamp & Content & \makecell[l]{Genl.} & LLaMA-2-7B & $\times$ & Token & $\times$ &   \textbf{G} \\ 

    \midrule

    MemPrompt~\citep{madaan2022memory} &   Content & \makecell[l]{Genl.} & GPT-3 & $\checkmark$ & Token & $\times$ &   \textbf{G} \\ 

    TeachMe~\citep{dalvi2022towards} &   Content & \makecell[l]{Ext., Genl.} & GPT-3 & $\checkmark$ & Token & $\times$ &   \textbf{G} \\ 

    MaLP~\citep{zhang2024llm} &   Dialogues & \makecell[l]{Abs., Genl.} & GPT-3.5, LLaMA-7B, LLaMA-13B & $\checkmark$ & Token & $\checkmark$ (LoRA) &   \textbf{G},   \textbf{C} \\

    MemoRAG~\citep{qian2024memorag} &   Dialogues & \makecell[l]{Ext., Genl.} & Qwen2-7B-Instruct, Mistral-7B-Instruct & $\checkmark$ & Token & $\times$ & \textbf{G} \\ 
    
    IPA~\citep{salemi2023lamp}~\lamp &   Content & \makecell[l]{Abs., Genl.} & Flan-T5-base & $\checkmark$ & Token & $\times$ & \textbf{G},   \textbf{C} \\ 

    FiD~\citep{salemi2023lamp}~\lamp &   Content & \makecell[l]{Abs., Genl.} & Flan-T5-base & $\checkmark$ & Token & $\checkmark$ & \textbf{G},   \textbf{C} \\ 

    MSP~\citep{zhong2022less} &   Dialogues & \makecell[l]{Genl.} & DialoGPT & $\checkmark$ & Token & $\times$ & \textbf{G} \\ 

    AuthorPred~\citep{li2023teach} &   Content & \makecell[l]{Abs., Genl.} & T5-11B & $\checkmark$ & Token & $\checkmark$ & \textbf{G},   \textbf{C} \\ 

    PEARL~\citep{mysore2023pearl} &   Content & \makecell[l]{Genl.} & Davinci-003, GPT-3.5-Turbo & $\checkmark$ & Token & $\times$ & \textbf{G} \\ 

    ROPG~\citep{salemi2024optimization}~\lamp &   Content & \makecell[l]{Abs., Genl.} &  Flan-T5-XXL-11B & $\checkmark$ & Token & $\times$ & \textbf{G},   \textbf{C} \\ 

    HYDRA~\citep{zhuang2024hydra}~\lamp &   Content & \makecell[l]{Abs., Genl.} & GPT-3.5-Turbo  & $\checkmark$ & Token & (Adaptor) & \textbf{G},   \textbf{C} \\ 

    PRM~\citep{zhuang2024hydra}~\lamp &  Content & \makecell[l]{Abs., Genl.} & GPT-4o-mini  & $\checkmark$ & Token & $\times$ & \textbf{G}, \textbf{C}    \\ 

    PersonaAgent~\citep{zhang2025personaagent}~\lamp &  Content & \makecell[l]{Abs., Genl.} & Claude-3.5-Sonnet  & $\checkmark$ & Token & $\times$ & \textbf{C}    \\

    \midrule

    UEM~\citep{doddapaneni2024user} &   Interactions & \makecell[l]{Abs.} & Flan-T5-base, Flan-T5-large & $\times$ & Embedding & $\checkmark$ &   \textbf{C} \\ 

    PERSOMA~\citep{hebert2024persoma} &   Interactions & \makecell[l]{Genl.} & PaLM-2 & $\times$ & Embedding & $\checkmark$ (LoRA) & \textbf{G} \\ 

    REGEN~\citep{sayana2024beyond} &   Interactions & \makecell[l]{Genl.} & PaLM-2 & $\times$ & Embedding & $\times$ & \textbf{G} \\ 

    PeaPOD~\citep{ramos2024preference} &   Interactions & \makecell[l]{Genl.} & T5-small & $\times$ & Embedding & $\checkmark$ & \textbf{G},   \textbf{R} \\ 

    ComMer~\citep{zeldes2025commer}~\lamp \perltqa & Content & \makecell[l]{Genl.} & Gemma-2B & $\times$ & Embedding & $\times$ & \textbf{G} \\ 

    PPlug~\citep{liu2024llms+}~\lamp &   Content & \makecell[l]{Abs., Genl.} & Flan-T5-XXL-11B & $\times$ & Embedding & $\times$ &   \textbf{G},   \textbf{C} \\ 
    
    User-LLM~\citep{ning2024user} &   Interactions & \makecell[l]{Genl.} & PaLM-2-XXS & $\times$ & Embedding & $\checkmark$ & \makecell[l]{  \textbf{G},   \textbf{R}} \\ 

    RECAP~\citep{liu2023recap} &   Dialogues & \makecell[l]{Genl.} & DialoGPT & $\checkmark$ & Embedding & $\checkmark$ &   \textbf{G} \\ 

    GSMN~\citep{wu2021personalized} &  \makecell[l]{User profile \\ Comment} & \makecell[l]{Genl.} & DialoGPT & $\checkmark$ & Embedding & $\checkmark$ &   \textbf{G} \\ 

    \midrule[1.2pt]
    \rowcolor{gray!10}  
    \multicolumn{8}{l}{\color{black}\textsection\ref{sec: personalized adatation} \quad \textbf{Personalized Adaptation}} \\ 
    \midrule

     PLoRA~\citep{DBLP:conf/aaai/ZhangWYXZ24} & User profile (ID) & \makecell[l]{Abs.} & BERT, RoBERTa, Flan-T5 & $\times$ & $\times$ & LoRA &   \textbf{C}\\
    UserIdentifier~\citep{mireshghallah2021useridentifier} & User profile & \makecell[l]{Abs.} & RoBERTa-base & $\times$ & $\times$ & - &   \textbf{C} \\
    
     LM-P~\citep{wozniak2024personalized} & User profile (ID) & \makecell[l]{Abs., Genl.} & \makecell[l]{Mistral-7B, Flan-T5, Phi-2, \\ StableLM, GPT-3.5, GPT-4} & $\times$ & $\times$ & LoRA &   \textbf{G},   \textbf{C} \\  

     Review-LLM~\citep{peng2024llm} &   Interactions & \makecell[l]{Genl.} & GPT-3.5-Turbo, GPT-4o, LLaMA-3-8B & $\times$ & $\checkmark$ & LoRA &   \textbf{G}\\

     MiLP~\citep{zhang2024personalized} & \makecell[l]{  Content  \\   Dialogues} & \makecell[l]{Genl.}  & \makecell[l]{DialoGPT, RoBERTa, \\  LLaMA-2-7B, LLaMA-2-13B} & $\times$ & $\times$ & LoRA &   \textbf{G}\\

    RecLoRA~\citep{zhu2024lifelong} &   Interactions & \makecell[l]{Genl.} &  Vicuna-7B  & $\checkmark$ & $\checkmark$ & LoRA &   \textbf{R}\\

    iLoRA~\citep{kongcustomizing} &   Interactions & \makecell[l]{Genl.} &   LLaMA-2-7B  & $\times$ & $\times$ & LoRA &   \textbf{R}\\

    \hline

    UserAdapter~\citep{zhong2021useradapter} &   Interactions & \makecell[l]{Abs., Genl.} & RoBERTa-base & $\times$ & $\times$ & Prefix-tuning &   \textbf{R}\\
    
    PocketLLM~\citep{peng2024pocketllm} & Content & - & RoBERTa-large, OPT-1.3B & $\times$ & $\times$ & MeZo &   \textbf{C}\\

    OPPU~\citep{DBLP:conf/emnlp/Tan000Y024}~\lamp  &   Content & \makecell[l]{Abs., Genl.} & LLaMA-2-7B &  (o) & (o) & LoRA &   \textbf{G},   \textbf{C} \\

    CoPe~\citep{bu2025personalized}~\lamp \longlamp  &   Content & \makecell[l]{Genl.} & Mistral-7B-Instruct-v0.3 & $\times$ & $\times$ & LoRA &   \textbf{G}\\

    PER-PCS~\citep{tan2024personalized}~\lamp &   Content & \makecell[l]{Abs., Genl.} & LLaMA-2-7B & (o) & (o) & LoRA &   \textbf{G},   \textbf{C} \\ 

    PROPER~\citep{zhang2025proper}~\lamp &   Content & \makecell[l]{Abs., Genl.} & LLaMA-2-7B & $\times$ & $\times$ & LoRA &   \textbf{G}, \textbf{C} \\

     \citep{wagner2024personalized} &   Content & \makecell[l]{Abs.} & GPT-2 & $\times$ & $\times$ & LoRA &   \textbf{C} \\ 

     FDLoRA~\citep{qi2024fdlora} &   Content & - & LLaMA-2-7B  & $\times$ & $\times$ & LoRA &   \textbf{C} \\ 

    \midrule[1.2pt]
    \rowcolor{gray!10}  
    \multicolumn{8}{l}{\color{black}\textsection\ref{sec: personalized alignment} \quad \textbf{Personalized Alignment}} \\
    
    \midrule
    \cite{wu2024aligning} &   Dialogues & \makecell[l]{Genl.}  & \makecell[l]{Qwen2-7B-Instruct, LLaMA-3-8B-Instruct, \\ Mistral-7B-Instruct-v0.3} & $\times$ & $\times$ &$\checkmark$ &   \textbf{G} \\
    PLUM~\citep{magister2024way}  &   Dialogues & \makecell[l]{Genl.}  & LLaMA-3-8B-Instruct & $\times$ & $\times$ &LoRA &   \textbf{G} \\
    \cite{lee2024aligning}& User Profile & \makecell[l]{Genl.} & Mistral-7B-v0.2 & $\times$ & $\times$ &$\checkmark$ &   \textbf{G} \\

    \midrule
    
    MORLHF~\citep{wu2023fine}  &    Preference &  \makecell[l]{Genl.} & GPT-2, T5-large & $\times$ & $\times$ &$\checkmark$ &   \textbf{G} \\
    MODPO~\citep{zhou2023beyond}  &    Preference & \makecell[l]{Genl.} & LLaMA-7B & $\times$ & $\times$ &LoRA &   \textbf{G} \\
    Personalized Soups~\citep{jang2023personalized}  &    Preference & \makecell[l]{Genl.} & Tulu-7B & $\times$ & $\times$ &LoRA &   \textbf{G} \\
    Reward Soups~\citep{rame2024rewarded}  &    Preference & \makecell[l]{Genl.} & LLaMA-7B & $\times$ & $\times$ &LoRA &   \textbf{G} \\
    MOD~\citep{shi2024decoding}  &    Preference  & \makecell[l]{Genl.} & LLaMA-2-7B & $\times$ & $\times$ &$\checkmark$ & 
      \textbf{G} \\
    PAD~\citep{chen2024pad}  &    Preference & \makecell[l]{Genl.} & LLaMA-3-8B-Instruct, Mistral-7B-Instruct & $\times$ & $\times$ &LoRA &   \textbf{G} \\
    PPT~\citep{lau2024personalized}  &    Preference & \makecell[l]{Genl.} & Self-defined & $\times$ & $\times$ &$\checkmark$ &   \textbf{G} \\
    VPL~\citep{poddar2024personalizing}  &    Preference & \makecell[l]{Genl.} & GPT-2, LLaMA-2-7B & $\times$ & $\times$ &LoRA &   \textbf{G} \\
    
\bottomrule[1.2pt]
\end{tabular}
}

\label{tab:methods}
\end{table*}

\section{Conclusions}

This survey offers a comprehensive overview of PLLMs, focusing on personalized responses to individual user data. It presents a taxonomy categorizing approaches into three key perspectives: Personalized Prompting (Input Level), Personalized Adaptation (Model Level), and Personalized Alignment (Objective Level), with further subdivisions.  A detailed method summarization is shown in Table~\ref{tab:methods}. We highlight current limitations and suggest future research directions, providing valuable insights to advance PLLM development. Beyond taxonomy, we articulate a vision for \emph{memory-centric PLLMs}, highlighting three desired capabilities—\emph{remember}, \emph{adapt}, and \emph{evolve}—and analyze their inherent trade-offs among performance, privacy, and efficiency. 
We further examine domain-specific challenges and discuss future directions. 
Together, these contributions \textbf{not only summarize the state of the art but also chart promising pathways for advancing PLLM research and practical deployment}.

\newpage
\bibliographystyle{unsrtnat}
\bibliography{references}  

\newpage
\appendix
\section{Notation Summary}
\vspace{-18pt}
\begin{table}[H]
\small
\centering
\label{tab:notations}
\renewcommand{\arraystretch}{1.6}  
\caption{Notation summary used throughout the paper. Symbols are grouped into \textit{Spaces}, \textit{Functions}, \textit{Models}, \textit{Operations}, and \textit{Variables}. This table consolidates symbols from the survey.} 
\label{tab:notation}
\rowcolors{2}{white}{mygrey!15}
\begin{tabular}{cl}
\toprule[1.3pt]
\textbf{Symbol} & \textbf{Description} \\

\midrule[1.2pt]
\rowcolor{white}\multicolumn{2}{l}{\textit{• Spaces}} \\
\hline
$\mathcal{C}$ & Space of all personalized data \\
$\mathcal{Z}$ & Latent abstraction space (preferences, profiles, tendencies) \\
$\mathcal{H}$ & Personalization state space (style, policy, reasoning priors, etc.) \\
$\mathcal{S} = \{d_k\}_{k=1}^K$ & Predefined human preference dimensions (e.g., ``helpfulness'') \\
$Q$ & Query space \\
$Y$ & Output / response space \\
$\mathcal{K}$ & General or world knowledge available to the LLM \\
$U = \{u_i\}_{i=1}^N$ & Set of users \\

\midrule[1.2pt]
\rowcolor{white}\multicolumn{2}{l}{\textit{• Functions / Mappings}} \\
\hline
$\phi: \mathcal{C} \to \mathcal{Z}$ & Abstraction / summarization mapping \\
$\psi: (\mathcal{C} \times Q) \to \mathcal{H}$ & Personalization encoder (possibly query-aware) \\
$f_{\mathrm{ext}}: Q \times \mathcal{C} \to Y$ & Extraction operator (factual lookup) \\
$f_{\mathrm{abs}}: Q \times \mathcal{Z} \to Y$ & Abstraction operator (summary-based answering) \\
$p_\theta(y \mid q, h, \mathcal{K})$ & Conditional distribution for generalization queries \\

\midrule[1.2pt]
\rowcolor{white}\multicolumn{2}{l}{\textit{• Models}} \\
\hline
$M_0: Q \to Y$ & Base large language model (non-personalized) \\
$M_i = P(M_0, C_i; \theta)$ & Personalized model for user $u_i$ \\
$P$ & Personalization operator (e.g., prompting, adaptation, alignment) \\
$\theta$ & Tunable parameters (within LLM or added modules, e.g., PEFT/LoRA) \\

\midrule[1.2pt]
\rowcolor{white}\multicolumn{2}{l}{\textit{• Operations / Sets}} \\
\hline
\rowcolor{mygrey!15} $\Sigma(C_i)$ & Explicit factual tuples contained in $C_i$ (for Extraction) \\

\midrule[1.2pt]
\rowcolor{white}\multicolumn{2}{l}{\textit{• Variables}} \\
\hline
$u_i \in U$ & The $i$-th user \\
$C_i \subset \mathcal{C}$ & Personalized data of user $u_i$ \\
$Q_i = \{q^{(i)}_j\}_{j=1}^{n_i}$ & Query set of user $u_i$ \\
$q^{(i)}_j \in Q$ & The $j$-th query of user $u_i$ \\
$y^{(i)}_j \in Y$ & Predicted response for $q^{(i)}_j$ (model output) \\
$\hat{y}^{(i)}_j \in Y$ & Desired personalized output for $q^{(i)}_j$ (ground truth) \\
$(c^{(i)}_j, r^{(i)}_j) \in C_i$ & Training pair: context/content and reference response \\
$r^{(i)}_j \in Y$ & Reference response associated with $c^{(i)}_j$ in training data \\
$z_i = \phi(C_i)$ & Abstracted profile of user $u_i$ \\
$h_{i,j} = \psi(C_i, q^{(i)}_j)$ & Personalized state for user $u_i$ under query $q^{(i)}_j$ \\

\bottomrule[1.3pt]
\end{tabular}
\end{table}

\end{document}